\documentclass[english,twocolumn,transaction]{IEEEtran}
\usepackage[T1]{fontenc}
\usepackage{float}
\usepackage{amsmath}
\allowdisplaybreaks[4]
\usepackage{babel}
\usepackage{amsmath}
\usepackage{amssymb}
\usepackage{graphicx}
\usepackage{array}
\usepackage{subfigure}
\usepackage{cancel}
\usepackage{float}
\usepackage{amsthm}
\usepackage{amsmath}
\usepackage{array}
\usepackage{amssymb}
\usepackage{stackrel}
\usepackage{multirow}
\usepackage{colortbl}
\usepackage{tabu}
\usepackage{setspace}
\usepackage{url}%%in order to show the web link in the reference
\usepackage{algorithm}
\usepackage{algorithmic}
\usepackage{booktabs}
\usepackage{cite}
\usepackage{color}
\usepackage{subfigure}
\newcolumntype{Y}{>{\centering\arraybackslash}X}

\newcommand{\tabincell}[2]{\begin{tabular}{@{}#1@{}}#2\end{tabular}}
\makeatletter

\floatstyle{ruled}
\newfloat{algorithm}{tbp}{loa}
\providecommand{\algorithmname}{Algorithm}
\floatname{algorithm}{\protect\algorithmname}

\theoremstyle{plain}

\theoremstyle{definition}

\theoremstyle{plain}

\theoremstyle{plain}

\definecolor{mygray}{gray}{.9}
\definecolor{mygray1}{gray}{1.}
\definecolor{mypink}{rgb}{.99,.91,.95}
\definecolor{mycyan}{cmyk}{.3,0,0,0}

\IEEEoverridecommandlockouts
\usepackage{ifpdf}
\usepackage{cite}
\ifCLASSOPTIONcompsoc
\usepackage[caption=false,font=normalsize,labelfont=sf,textfont=sf]{subfig}
\else
\usepackage[caption=false,font=footnotesize]{subfig}
\fi
\usepackage{amsmath}
\@ifundefined{showcaptionsetup}{}{%
 \PassOptionsToPackage{caption=false}{subfig}}
\usepackage{subfig}
\makeatother

\usepackage{babel}
\newtheorem{define}{Definition}

\newtheorem{Trick}{Improvement}
\newtheorem{theo}{Theorem}

\renewcommand\figurename{Fig.}

\begin{document}
\title{Joint Resource Management for MC-NOMA: A Deep Reinforcement Learning Approach}
\author{Shaoyang Wang, Tiejun Lv,~\emph{Senior Member, IEEE},
Wei Ni,~\emph{Senior Member, IEEE}, Norman C. Beaulieu,~\IEEEmembership{Fellow,~IEEE},
 and Y. Jay Guo,~\IEEEmembership{Fellow,~IEEE}

\thanks{Manuscript received March 31, 2020; revised September 25, 2020 and October 30, 2020; accepted March 20, 2021.
The financial support of the BUPT Excellent Ph.D. Students Foundation (Grant No. CX2019106) is gratefully acknowledged. \emph{(Corresponding author: Tiejun Lv.)}

 S. Wang and T. Lv are with the Key Laboratory of Trustworthy Distributed
Computing and Service, Ministry of Education, and the School of Information and Communication Engineering, Beijing University of Posts and Telecommunications, Beijing 100876, China (e-mail: \{shaoyangwang, lvtiejun\}@bupt.edu.cn).

 W. Ni is with the Commonwealth Scientific and Industrial Research Organisation (CSIRO), Sydney 2122, Australia (e-mail: Wei.Ni@data61.csiro.au).

 N. C. Beaulieu is with the Beijing Key Laboratory for Network System Architecture and Convergence, and the School of Information and Communication Engineering, Beijing University of Posts and Telecommunications, Beijing 100876, China (e-mail: nborm@bupt.edu.cn).

 J. Guo is with the Global Big Data Technologies Centre, University of Technology Sydney, Ultimo, NSW 2007, Australia (e-mail: jay.guo@uts.edu.au).
}}
\maketitle
\begin{abstract}
This paper presents a novel and effective deep reinforcement learning (DRL)-based approach to addressing joint resource management (JRM) in a practical multi-carrier non-orthogonal multiple access (MC-NOMA) system, where hardware sensitivity and imperfect successive interference cancellation (SIC) are considered. We first formulate the JRM problem to maximize the weighted-sum system throughput. Then, the JRM problem is decoupled into two iterative subtasks: subcarrier assignment (SA, including user grouping) and power allocation (PA). Each subtask is a sequential decision process. Invoking a deep deterministic policy gradient algorithm, our proposed DRL-based JRM (DRL-JRM) approach jointly performs the two subtasks, where the optimization objective and constraints of the subtasks are addressed by a new \emph{joint reward} and \emph{internal reward} mechanism. A multi-agent structure and a convolutional neural network are adopted to reduce the complexity of the PA subtask. We also tailor the neural network structure for the stability and convergence of DRL-JRM. Corroborated by extensive experiments, the proposed DRL-JRM scheme is superior to existing alternatives in terms of system throughput and resistance to interference, especially in the presence of many users and strong inter-cell interference. DRL-JRM can flexibly meet individual service requirements of users.
\end{abstract}
\section{Introduction}
\renewcommand\figurename{Fig.}
Owing to enhanced multi-user diversity and high flexibility in resource allocation, multi-carrier (MC) multi-access techniques have been widely applied in wireless systems\cite{7263349}. Recently, non-orthogonal multi-access (NOMA) has attracted significant attention. It can improve spectral efficiency, by adopting superposition coding at the transmitter and successive interference cancellation (SIC) at the receiver. Combining MC and NOMA techniques, MC-NOMA extends the technical legacy of orthogonal frequency-division multi-access \cite{7676258}.

The design and deployment of resource management play a crucial role in \emph{power-domain MC-NOMA} (or ``MC-NOMA'' for short) systems \cite{6730679}. The resource management involves user grouping, subcarrier scheduling, and power allocation. Generally, the optimal joint resource management (JRM) algorithms require solving mixed integer nonlinear programming problems. Because MC-NOMA systems involve a much larger number of discrete and/or continuous optimization variables, most existing solvers become computationally prohibitive.

For MC-NOMA with perfect SIC, a variety of optimal and suboptimal JRM algorithms have been proposed, typically using optimization theory. Lei \emph{et al.} \cite{7587811} employed Lagrangian duality and dynamic programming to obtain suboptimal solutions for power and channel allocation in a downlink (DL) MC-NOMA system. For spectrum and energy efficient resource allocation, Song \emph{et al.} \cite{8403100} formulated a multi-objective optimization problem solved by convex programming. Fu \emph{et al.} \cite{8489876} also utilized convex programming and heuristic greedy algorithms to solve the subcarrier and power allocation problem, and designed a three-step resource allocation framework. Di \emph{et al.} \cite{7560605} investigated the power allocation and user scheduling to maximize the weighted-sum rate by designing a matching game. Zheng \emph{et al.} \cite{8375659} considered an uplink (UL) MC-NOMA system, and designed a Nash bargaining game for power allocation and user clustering. For the simpler UL power allocation problem, Fang \emph{et al.} \cite{8680645} used Lagrangian dual decomposition and Dinkelbach algorithm to derive closed-form expressions for the optimal power allocation. Moreover, Sun \emph{et al.} \cite{7812683} studied the joint power and subcarrier allocation in a full-duplex MC-NOMA system, and a successive convex approximation (SCA)-based suboptimal iterative scheme was presented. All the above studies assumed perfect SIC.

In practice, MC-NOMA is susceptible to imperfect SIC and subsequent error propagation. It is of significance to study the JRM of MC-NOMA systems in the presence of imperfect SIC (also known as imperfect NOMA). Wei \emph{et al.} \cite{7934461} considered an MC-NOMA system under imperfect channel state information CSI, and employed the branch-and-bound (B\&B) and difference of convex (DC) programming to maximize the power efficiency of the system. Cheng \emph{et al.} \cite{8538600} modeled the channel estimation error as a complex Gaussian random variable and proposed a low-complexity user scheduling and power allocation algorithm in a DL MC-NOMA system. The results of \cite{7934461} and \cite{8538600} are only applicable in the case with at most two users multiplexed per subcarrier. To reduce the power consumption of a delay-sensitive UL transmission, Xu \emph{et al.} \cite{xu2019uplink} presented an approximate algorithm to minimize the transmit power of imperfect NOMA. Zamani \emph{et al.} \cite{8540884} formulated the power allocation into a fractional programming problem, and obtained the optimal solution by using the Karush-Kuhn-Tucker conditions and DC programming. While power allocation has been studied in \cite{xu2019uplink} and \cite{8540884}, user grouping and subcarrier scheduling have not been well investigated. Adopting heuristics user scheduling, the authors of \cite{8755843} focused on the power allocation under imperfect SIC with no consideration of the quality of service (QoS). Cellik \emph{et al.} \cite{8269139} considered the power disparity and sensitivity of the SIC receiver, and formulated a joint cluster formation and resource allocation problem. The subproblem of cluster formation was solved by the blossom algorithm. The subproblem of resource allocation was solved by geometric programming. The focus of \cite{8269139} was on the UL, which however is distinctively different from the DL NOMA systems considered in this paper.

The above existing works typically have the following two limitations: 1) deriving optimal JRM schemes is intractable for MC-NOMA under imperfect SIC, and 2) assuming that the same number of users are multiplexed on each subcarrier. In addition, the operation of an SIC receiver requires that the power difference between the signal and noise to exceed a threshold (referred to as ``hardware sensitivity requirement'', which has been overlooked in the existing literature).

Reinforcement learning (RL) allows an agent to maximize a long-term discount reward and derive a solution on its own \cite{8714026}. Yang \emph{et al.} \cite{yang2019cache} utilized Q-learning to design a NOMA-based mobile edge computing framework. By integrating deep learning into RL, deep RL (DRL) addresses the challenges of Q-learning in the storage and look-up of the Q table. Yang \emph{et al.} \cite{8770605} employed a deep Q-network (DQN) to model the offloading problem for multi-user NOMA. Doan \emph{et al.} \cite{8869815} applied DRL to implement the power allocation in cache-aided NOMA systems. By employing the actor-critic RL, Zhang \emph{et al.} \cite{8641248} proposed a dynamic power allocation scheme. Zhang {et al.} \cite{8927898} and Giang \emph{et al.} \cite{giang2020hybrid} used DRL to obtain suboptimal solutions to the power allocation of UL MC-NOMA systems.  He \emph{et al.} \cite{8790780} used a DRL framework solve to the joint power allocation and channel assignment problem in a perfect two-user NOMA system. An attention-based neural network was applied to capture the sequential relations between the input and output of channel assignment problem. In our recent works, we studied multi-channel access in fast-changing channels \cite{9037116}, joint virtual network function placement-and-routing \cite{9024688}, and user grouping of NOMA systems \cite{8757016}, by utilizing a new policy gradient-based DRL technique. In this paper, we are interested in a different problem of joint subcarrier assignment and  power allocation in an MC-NOMA system with imperfect SIC. Although the above existing studies have revealed the potential of RL/DRL in the resource management, there is still paucity of research for applying DRL to the JRM of MC-NOMA, especially in the presence of hardware sensitivity requirements and imperfect SIC.

This paper optimizes the subcarrier assignment and power allocation of a DL MC-NOMA system, under imperfect channel station information (CSI), non-negligible SIC errors, and given hardware sensitivity. These practical factors have yet to be considered in the literature. We propose a novel DRL-based JRM (DRL-JRM) framework with the following contributions:
\begin{itemize}
\item A practical DL MC-NOMA system is studied, where: 1) the number of users multiplexed on different subcarriers can differ; 2) the hardware sensitivity and imperfect SIC cannot be overlooked; and 3) subcarrier assignment (including user grouping) and power allocation are jointly optimized. We maximize the weighted-sum throughput of the DL MC-NOMA system, by decoupling the JRM of the system into two RL subtasks.
\item A novel DRL-RM framework is designed for the two RL subtasks, where there are two modules: an \emph{SA module} responsible for subcarrier assignment and a \emph{PA module} responsible for power allocation. Both the SA and PA modules are implemented with a deep deterministic policy gradient (DDPG) algorithm. The optimization objective and constraints of the two subtasks are constructed as a new \emph{joint reward} and \emph{internal reward} mechanism.
\item A novel \emph{centralized action-value function} is designed to measure the reward in the PA module. In the \emph{centralized action-value function}, we employ a convolutional neural network (CNN) to guarantee that an agent can effectively utilize its own information by condensing the information of other agents into lower dimensions. Several new designs on the neural network are also proposed.
\end{itemize}

Extensive experiments verify that the proposed DRL-JRM scheme provides close-to-optimal results in the case of small-scale problems. In the case of large-scale problems, DRL-JRM is superior to existing benchmarks in terms of weighted-sum system throughput and resistance to interference.

The rest of this paper is organized as follows. In Section \uppercase\expandafter{\romannumeral2}, we present the system model and formulate the JRM problem. The problem is transformed into an RL task in Section \uppercase\expandafter{\romannumeral3}. In Section \uppercase\expandafter{\romannumeral4}, the new DRL-JRM framework is developed. Experiments are shown in Section \uppercase\expandafter{\romannumeral5}, followed by conclusions in Section \uppercase\expandafter{\romannumeral6}. Notations used are collected in Table \uppercase\expandafter{\romannumeral1}.
\section{Problem Statement}
\subsection{System Model}
Consider a classical cellular DL MC-NOMA system, where a base station (BS) serves $M$ users. The frequency band is $W$ (Hertz) with $N_{\mathrm{F}}$ orthogonal subcarriers.
\begin{table*}[htb]
	\centering{}
	\textbf{Table \uppercase\expandafter{\romannumeral1}}~~The key notations used in this paper.\\
\small
\setlength{\tabcolsep}{0.150mm}{
		\begin{tabular}{|c|l|c|l|}
        \hline
        \multicolumn{2}{|>{\columncolor{mygray}}c|}{\textbf{Section II}} & $W$ & bandwidth\\
        \hline
        $M$ & the number of users & $N_{\mathrm{F}}$ & number of subcarriers\\
        \hline
        $J_{i}$ & number of users multiplexed on subcarrier $i$ & $N_{\mathrm{max}}$ & maximum number of users per subcarrier\\
       \hline
        $h_{i,j}$ & the channel gain of the $j$-th user on subcarrier $i$ & $\tilde{h}_{i,m}$ & channel gain of user $m$ on subcarrier $i$\\
       \hline
       $\vartheta_{i,j}^{m}$ & indicator of whether the $j$-th user on subcarrier $i$ is user $m$ & $p_{\Delta}$ & hardware sensitivity requirement\\
       \hline
       $p_{i,j}$ & the power allocated to the $j$-th user on subcarrier $i$ &$a_{i,j}$ & symbol of the $j$-th user on subcarrier $i$\\
       \hline
       $\varepsilon$ & average estimation error of modulated symbol &$R_{i,j}$ & achievable rate the $j$-th user on subcarrier $i$\\
       \hline
        $\tilde{R}_{i,m}$ & the achievable rate the user $m$ on subcarrier $i$ & $R_{m}^{\mathrm{min}}$ & minimum data rate requested by user $m$\\
          \hline
        $\varpi_{m}$ & the priority weight of user $m$ & $d_{m}$ & distance form user $m$ to the BS\\
       \hline
       \hline
       \multicolumn{2}{|>{\columncolor{mygray}}c|}{\textbf{Section III}} & $\mathbf{a}_{t}^{\mathrm{u}}$ & the SA action at the decision step $t$\\
        \hline
        $T^{\mathrm{PA}}_{\mathrm{max}}$ & the maximum number of steps for the PA subtask & $\mathbf{a}_{m}^{\mathrm{u}}$ & the final SA action of agent $m$\\
       \hline
       $\mathbf{a}_{m,t}^{\mathrm{p}}$ & the PA action of user $m$ at the decision step $t$ & $s_{t}^{\mathrm{u}}$ & the SA state at the decision step $t$\\
       \hline
       $o_{i,m}$ & indicator of whether user $m$ is assigned to subcarrier $i$ & $s_{m,t}^{\mathrm{p}}$ & the PA state of user $m$ at the decision step $t$\\
       \hline
       $s_{\mathrm{self},m,t}^{\mathrm{p}}$ & the self-state of agent $m$ at the decision step $t$ &$\gamma$ & discounted factor of reward \\
       \hline
       $s_{\mathrm{other},m,t}^{\mathrm{p}}$ & state of the other agents for agent $m$ at the decision step $t$ & $r_{t}^{\mathrm{u,int}}$ & the internal reward of SA subtask\\
       \hline
       $Q(s_{t},a_{t})$ & the action-value function & $r_{t}^{\mathrm{u,jo}}$ & the joint reward of SA subtask\\
       \hline
       $r_{m,t}^{\mathrm{p,int}}$ & the internal reward of SA subtask for agent $m$ & $r_{m,t}^{\mathrm{p,jo}}$ & the joint reward of SA subtask for agent $m$\\
       \hline
	\end{tabular}}
\end{table*}
The BS and the users are all equipped with a single antenna\footnote{This paper studies the classical single-cell multi-user scenario, and the research under the multi-cell multi-user scenario will be our next work.}. In this paper, user $m$ ($m\in\left\{ 1,\ldots,M\right\}$) may occupy multiple subcarriers, and subcarrier $i$ ($i\in\left\{ 1,\ldots,N_{\mathrm{F}}\right\}$) may be occupied by multiple users. Let $J_{i}$ denote the number of users multiplexed on subcarrier $i$ and $N_{\mathrm{max}}$ be the maximum number of users per subcarrier. Different from most existing studies which assumed at most two users per subcarrier, we allow more and different numbers of users to be assigned per subcarrier\footnote{The complexity of the SIC decoding at the users grows linearly with the number of users (or in other words, streams) multiplexed per subcarrier. Such (linear) complexity is in general scalable.}.
\begin{figure}[htb]
\centering{}\includegraphics[scale=0.5]{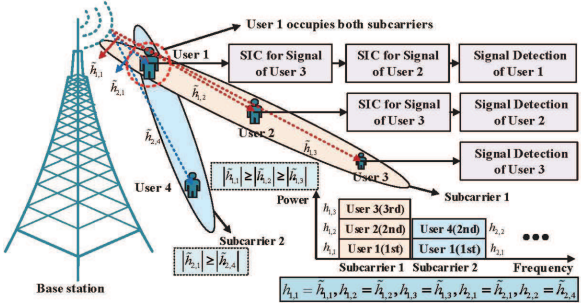}
\caption{An simple example of our MC-NOMA system. Four users are multiplexed on two subcarriers, where users 1, 2 and 3 are multiplexed on subcarrier 1; and user 1 and user 4 are multiplexed on subcarrier 2. }\label{system_model}
\end{figure}

Assume that the channel gains of the users on subcarrier $i$ satisfy: $|h_{i,1}|\geq\ldots\geq| h_{i,j}|\geq\ldots\geq|h_{i,J_{i}}|$, where $h_{i,j}$ is the channel gain of the $j$-th user on subcarrier $i$. Let a binary variable $\vartheta_{i,j}^{m}\in\left\{ 0,1\right\}$ indicate whether the $j$-th user multiplexed at subcarrier $i$ is user $m$. If the $j$-th user ($j\in\left\{ 1,\ldots,J_{i}\right\}$) on subcarrier $i$ is user $m\in\left\{ 1,\ldots,M\right\}$, $\vartheta_{i,j}^{m}=1$; otherwise, $\vartheta_{i,j}^{m}=0$. So, $\sum_{m=1}^{M}\sum_{j=1}^{J_{i}}\vartheta_{i,j}^{m}=J_{i},\forall i$. The transmit signal of the BS at the $i$-th subcarrier is given by
\begin{align}
x_{i}=\stackrel[m=1]{M}{\sum}\stackrel[j=1]{J_{i}}{\sum}\left(\vartheta_{i,j}^{m}\sqrt{p_{i,j}}a_{i,j}\right),\forall i
\end{align}
where $a_{i,j}\in\mathbb{C}$ is the modulated symbol of the $j$-th user on subcarrier $i$ and $\mathbb{E}\left[\mid a_{i,j}\mid^{2}\right]=1$. $p_{i,j}$ is the power allocated to the $j$-th user on subcarrier $i$. The received signal of user $m$ at the $i$-th subcarrier is given by
\begin{align}
y_{i,m}=\tilde{h}_{i,m}\stackrel[m=1]{M}{\sum}\stackrel[j=1]{J_{i}}{\sum}\left(\vartheta_{i,j}^{m}\sqrt{p_{i,j}}a_{i,j}\right)+z_{i,m}
\end{align}
where $z_{i,j}\in\mathbb{C}$ is the additive white Gaussian noise at the $j$-th user on subcarrier $i$, and $z_{i,j}\sim\mathcal{CN}\left(0,\sigma_{i,j}^{2}\right)$.  $\tilde{h}_{i,m}\!=\!\frac{g_{i,m}}{\mathrm{PL}_{m}}\in\mathbb{C}$ is the channel gain of user $m$ on subcarrier $i$, accounting for both the path loss ($\mathrm{PL}_{m}$) and small-scale fading ($g_{i,m}\sim\mathcal{CN}\left(0,1\right)$) \cite{7934461}. As show in Fig. \ref{system_model},
\begin{align}\label{transform}
\tilde{h}_{i,m}=\sum_{j=1}^{J_{i}}\vartheta_{i,j}^{m}h_{i,j}.
\end{align}

The SIC receiver first detects the strongest interference, and subtracts it from the received signal, then the second strongest, so on and so forth, until the user detects its intended signal. Generally, the operation of the receiver requires that the difference between the signal power and the noise power exceeds a threshold, depending on the hardware sensitivity, referred to as \emph{power disparity and sensitivity constraint} (PDSC) \cite{8254637}.
\begin{define}
For the $j$-th user on subcarrier $i$, PDSC is
\begin{align}
\mid h_{i,j}\mid^{2}\left(p_{i,j}-\stackrel[k=1]{j-1}{\sum}p_{i,k}\right)\geq p_{\Delta},\forall i,j
\end{align}
where $p_{\Delta}$ is a specific hardware sensitivity requirement.
\end{define}
Imperfect SIC suffers from residual interference after SIC, mainly due to imperfect amplitude and phase estimation \cite{xu2019uplink}. Define the original signal of the $j$-th user on subcarrier $i$ as $x_{i,j}=\sqrt{p_{i,j}}a_{i,j}$ and the estimated signal as $\hat{x}_{i,j}$. The residual interference $I_{i,j}^{\mathrm{re}}$ to the $j$-th user on subcarrier $i$ after SIC is
\begin{align}
\resizebox{.9\hsize}{!}{$I_{i,j}^{\mathrm{re}}=\stackrel[k=j+1]{J_{i}}{\sum}\mid h_{i,k}\mid^{2}\mid x_{i,k}-\hat{x}_{i,k}\mid^{2}=\stackrel[k=j+1]{J_{i}}{\sum}p_{i,k}\mid h_{i,k}\mid^{2}\mid a_{i,k}-\hat{a}_{i,k}\mid^{2}.
$}
\end{align}
Let $\ensuremath{\varepsilon_{i,k}\!=\!\mathbb{E}\{|a_{i,k}\!-\!\tilde{a}_{i,k}|^{2}\}}$ be the fractional error after cancelling the $k$-th user at subcarrier $i$. As suggested in \cite{1371734}, $\ensuremath{\varepsilon_{i,k}}$ can be approximated by a Gaussian distribution. Assume that the channel estimation errors are independent and identically distributed among different users at different subcarriers \cite{1427678}, and thus $\ensuremath{\varepsilon_{i,k}}\!=\!\ensuremath{\varepsilon}$, as assumed in \cite{8755843,8422363,8765793}. Hence, the residual error is $I_{i,j}^{\mathrm{re}}\!=\!\varepsilon^{2}\sum_{k=j+1}^{J_{i}}p_{i,k}\mid h_{i,k}\mid^{2}$.

The achievable rate $R_{i,j}$ of the $j$-th user on subcarrier $i$ is
\begin{align}
R_{i,j}=\frac{W}{N_{\mathrm{F}}}\log_{2}\left(1+\frac{p_{i,j}\mid h_{i,j}\mid^{2}}{\stackrel[k=1]{j-1}{\sum}p_{i,k}\mid h_{i,j}\mid^{2}+I_{i,j}^{\mathrm{re}}+\sigma_{i,j}^{2}}\right)
\end{align}
where $\sum^{j-1}_{k=1}p_{i,k}\mid h_{i,j}\mid^{2}$ is the interference from the users with lower powers than the $j$-th user on subcarrier $i$. Let $\tilde{R}_{i,m}$ be the achievable rate of user $m$ on subcarrier $i$. Then, $\tilde{R}_{i,m}=\sum_{j=1}^{J_{i}}\vartheta_{i,j}^{m}R_{i,j}$ by (\ref{transform}).
\subsection{Optimization Problem Formulation}
We aim to jointly optimize the subcarrier assignment and power allocation to maximize the weighted-sum throughput. The joint resource management can be formulated as:
\begin{align}\label{hard}
\underset{p_{i,j},\vartheta_{i,j}^{m}}{\mathrm{maximize}}\;&\stackrel[m=1]{M}{\sum}\varpi_{m}\left(\stackrel[i=1]{N_{\mathrm{F}}}{\sum}\stackrel[j=1]{J_{i}}{\sum}\vartheta_{i,j}^{m}R_{i,j}\right){} \tag{\emph{OP1}} \\ \nonumber
\mbox{s.t.} \quad
\mathrm{C}1:\:&\stackrel[m=1]{M}{\sum}\stackrel[i=1]{N_{\mathrm{F}}}{\sum}\stackrel[j=1]{J_{i}}{\sum}\vartheta_{i,j}^{m}p_{i,j}\leq P_{\mathrm{total}},\forall i,j,m,\\ \nonumber
 \mathrm{C}2:\:&|h_{i,j}|^{2}\left(\bar{p}_{i,j}-\stackrel[k=1]{j-1}{\sum}p_{i,k}\right)\geq p_{\Delta},\forall i,j,\\ \nonumber
 \mathrm{C}3:\:&\stackrel[j=1]{J_{i}}{\sum}\vartheta_{i,j}^{m}p_{i,j}\geq0,\forall i,j,m, \\ \nonumber \mathrm{C}4:\:&\stackrel[m=1]{M}{\sum}\stackrel[j=1]{J_{i}}{\sum}\vartheta_{i,j}^{m}\leq N_{\mathrm{max}},\\ \nonumber
 \mathrm{C}5:\:&\stackrel[i=1]{N_{\mathrm{F}}}{\sum}\stackrel[j=1]{J_{i}}{\sum}\vartheta_{i,j}^{m}R_{i,j}\geq R_{m}^{\mathrm{min}},\forall i,j,m, \\ \nonumber \mathrm{C}6:\:&\stackrel[j=1]{J_{i}}{\sum}\vartheta_{i,j}^{m}\in\left\{ 0,1\right\} ,\forall i,j,m,
\end{align}
where $0\!<\!\varpi_{m}\!<\!1$ is the weight of user $m$ accounting for its priority. As in \cite{7812683}, we set $\varpi_{m}=d_{m}/\left(\max_{i}\left(d_{i}\right)\right)$. $d_{m}$ is the distance from user $m$ to the BS. Constraint C1 specifies the maximum transmit power of the BS, $P_{\mathrm{total}}$. Constraint C2 ensures that the SIC receiver can successfully perform SIC. Constraint C3 ensures that non-negative powers. Constraint C4 limits the number of users per subcarrier. Constraint C5 specifies the minimum rate constraint of user $m$, $R_{m}^{\mathrm{min}}$, which is part of the QoS requirement of the user. Constraint C6 indicates that a user can only be allocated to a subcarrier once.
\begin{theo}
The optimization problem \emph{OP1} is NP-complete.
\end{theo}
\begin{IEEEproof}
The joint power and subcarrier allocation problem under perfect SIC is a special case (i.e., $\varepsilon^{2}\!=\!0$ and $p_{\Delta}\!=\!0$) of \emph{OP1}. The latter is proved to be an NP-complete problem \cite{7587811}. Thus, the NP-completeness of \emph{OP1} can be also proved.
\end{IEEEproof}
To solve \emph{OP1}, we design a novel DRL-JRM framework. Given the channel gain, QoS constraint and user priority, the framework, including subcarrier assignment and power allocation, is conducted at every slot.
\section{Transformation to RL Task}
\begin{figure}[h]
\centering{}\includegraphics[scale=0.58]{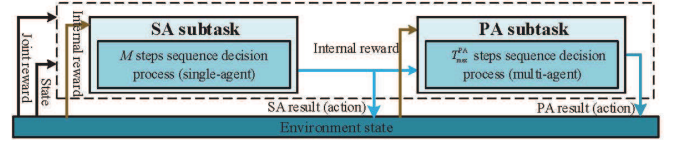}
\caption{Transformation of optimization problem \emph{OP1}.}\label{resource_stru_sample_new}
\end{figure}

In order to transform \emph{OP1} into an RL task, we decompose \emph{OP1} into two iterative subtasks which run in an alternating manner: an SA subtask and a PA subtask, as shown in Fig. \ref{resource_stru_sample_new}. The SA subtask is responsible for subcarrier assignment, and the PA subtask is responsible for power allocation. When solving the \emph{OP1}, the SA subtask is executed first and the SA result is obtained. Based on the SA result, the PA subtask is executed to obtain the PA result. The optimal solution is achieved after several iterations of the SA and PA subtasks.

To efficiently solve the SA and PA subtasks, RL is employed. A standard RL process is defined as a Markov decision process. At each decision step $t$, an agent observes state $s_{t}\in\mathcal{S}$, executes action $a_{t}\in\mathcal{A}$, and receives a scalar reward $r_{t}$. Based on the selected actions from the target policy $\pi$, the agent continuously interacts with a system environment $E$ to maximize the expected future rewards. The future discounted reward at step $t$ is $\mathcal{R}_{t}=\sum_{t^{'}=t}^{T}\gamma^{(t^{'}-t)}r_{t^{'}}$, where $T$ is the total number of steps of the RL task, and $\gamma$ is the discounted factor.

We transform the two subtasks into two sequential decision processes. The SA subtask is an $M$-step decision process. At each decision step, the agent outputs the SA result of a user. The process terminates after all users are assigned, and thus $M$ users need $M$ steps.  Different from the SA subtask, the PA subtask is a $T_{\mathrm{max}}^{\mathrm{PA}}$-step sequential decision process. At each step, the PA results of all users are output simultaneously. The outputs are the power change value of each user on each subcarrier, rather than the actual power value to be allocated. The process terminates after $T_{\mathrm{max}}^{\mathrm{PA}}$ steps.

Since each decision step in the PA subtask needs to output the PA results of all users at the same time, a multi-agent technology can be utilized. Each user corresponds to an agent, thus, a total of $M$ agents are needed.
\subsection{Action}
In the SA subtask, the action of each step is the SA result for a user. The action $\mathbf{a}_{t}^{\mathrm{u}}$ is defined as $\mathbf{a}_{t}^{\mathrm{u}}\!=\![ \pi_{1,t}^{\mathrm{u}},\ldots,\pi_{N_{\mathrm{F}},t}^{\mathrm{u}}]^{\mathrm{T}}$,  $\pi_{i,t}^{\mathrm{u}}\!\in\!\left\{ 0,1\right\} ,\forall i$. If subcarrier $i$ is assigned to the designated user of the $t$-th step, then $\pi_{i,t}^{\mathrm{u}}=1$; otherwise, $\pi_{i,t}^{\mathrm{u}}=0$.

In the PA subtask, the action of agent $m$ $\mathbf{a}_{m,t}^{\mathrm{p}}\!=\![\varrho_{1,m,t},\ldots,\varrho_{i,m,t},\ldots,\varrho_{N_{\mathrm{F}},m,t}]^{\mathrm{T}}
$, and $\varrho_{i,m,t}\!\in\!\left\{ -1,0,1\right\} ,\forall i$. $\varrho_{i,m,t}\!=\!1$ means power should be increased; $\varrho_{i,m,t}\!=\!0$ means power remains unchanged; $\varrho_{i,m,t}\!=\!-1$ means power should be reduced. Define $\vartheta$ as the magnitude of the power change at each step and $\mathbf{v}_{m,t}$ as the power indicator (rather than an actual power value). $\mathbf{v}_{m,t}=[v_{1,m,t},\ldots,v_{N_{\mathrm{F}},m,t}]^{\mathrm{T}}$, and $v_{i,m,t}$ evaluates the actual power value of user $m$ on subcarrier $i$ at the decision step $t$. Based on the PA action $\mathbf{a}_{m,t}^{\mathrm{p}}$, $\mathbf{v}_{m,t+1}$ in the next PA indicator state is $\mathbf{v}_{m,t+1}=\mathbf{v}_{m,t}+\vartheta\mathbf{a}_{m,t}^{\mathrm{p}}$. If $v_{i,m,t+1}<0$, then $v_{i,m,t+1}$ is reset to 0. The actual allocated power for user $m$ on subcarrier $i$ at step $t$ is given by
\begin{align}\label{allocated power}
 p_{i,m,t}=P_{t\mathrm{otal}}v_{i,m,t}/\left(\sum_{m=1}^{M}\sum_{i=1}^{N_{\mathrm{F}}}v_{i,m,t}\right).
\end{align}
\subsection{State}
In the SA subtask, $s_{t}^{\mathrm{u}}$ includes user priorities $\mathbf{W}$, QoS constraints $\mathbf{R}^{\mathrm{min}}$, channel gains $\tilde{\mathbf{H}}$ and the current state of subcarriers being occupied $\mathbf{O}_{t}$, i.e., $s_{t}^{\mathrm{u}}=\{ \mathbf{W},\mathbf{R}^{\mathrm{min}},\mathbf{H},\mathbf{O}_{t}\}$, where $\mathbf{W}=[\varpi_{1},\ldots,\varpi_{M}]$; $\mathbf{R}^{\mathrm{min}}=[R_{1}^{\mathrm{min}},\ldots,R_{M}^{\mathrm{min}}]$; $\tilde{\mathbf{H}}=[\tilde{\mathbf{h}}_{1},\ldots,\tilde{\mathbf{h}}_{M}]$, $\tilde{\mathbf{h}}_{m}=[\tilde{h}_{1,m},\ldots,\tilde{h}_{i,m},\ldots,\tilde{h}_{N_{\mathrm{F}},m}]^{\mathrm{T}}$; and $\mathbf{O}_{t}\!=\![\mathbf{o}_{1,t},\ldots,\mathbf{o}_{M,t}]$ and $\mathbf{o}_{m,t}\!=\![o_{1,m,t},\ldots,o_{i,m,t},\ldots,o_{N_{\mathrm{F}},m,t}]^{\mathrm{T}}$. $o_{i,m,t}\!\in\!\left\{ 0,1\right\}$ indicates whether user $m$ is assigned to subcarrier $i$ at step $t$. If user $m$ is assigned to the subcarrier $i$, then $o_{i,m,t}\!=\!1$; otherwise, $o_{i,m,t}\!=\!0$. Based on the definition of $\vartheta_{i,j}^{m}$, we have $o_{i,m}=\sum_{j=1}^{J_{i}}\vartheta_{i,j}^{m}$. In $s_{t}^{\mathrm{u}}$, all of $\mathbf{W}$, $\mathbf{R}^{\mathrm{min}}$ and $\tilde{\mathbf{H}}$ are fixed, while $\mathbf{O}_{t}$ changes with the SA action.

In the PA subtask, the $M$ users are $M$ agents. For each agent $m$, state $s_{m,t}^{\mathrm{p}}$ includes the information (e.g., user priority, QoS constraint, channel gain, and SA result) of other agents in addition to its own information. Thus, $\ensuremath{s_{m,t}^{\mathrm{p}}\!=\!\{\mathbf{s}_{\mathrm{self},m,t}^{\mathrm{p}},\mathbf{s}_{\mathrm{other},m,t}^{\mathrm{p}}\}}$, where $\mathbf{s}_{\mathrm{self},m,t}^{\mathrm{p}}\!=\![\varpi_{m},R_{m}^{\mathrm{min}},
(\mathbf{\tilde{h}}_{m})^{\mathrm{T}},\left(\mathbf{a}_{m}^{\mathrm{u}}\right)^{\mathrm{T}},\left(\mathbf{v}_{m,t}\right)^{\mathrm{T}}]
$, $\mathbf{s}_{\mathrm{other},m,t}^{\mathrm{p}}\\=
[(\mathbf{s}_{\mathrm{self},1,t}^{\mathrm{p}})^{\mathrm{T}},
\cdots,(\mathbf{s}_{\mathrm{self},k,t}^{\mathrm{p}})^{\mathrm{T}},\cdots,(\mathbf{s}_{\mathrm{self},M,t}^{\mathrm{p}})^{\mathrm{T}}]^{\mathrm{T}}
$, and $k\!\neq\!m$.
\subsection{Reward Function Design}
A new \emph{joint reward} and \emph{internal reward} mechanism is designed. The optimization objective is satisfied by a \emph{joint reward}, while constraints are satisfied by an \emph{internal reward}.

We refer to a complete SA and PA iteration as an ``epoch''. At the end of each epoch (iteration), we substitute the SA and PA results in \emph{OP1}, evaluate the objective value, and use it as the joint reward. The internal reward is the reward of environmental feedback when a non-suitable SA result (non-SSAR, not satisfy constraints C4 and C6) or non-suitable PA result (non-SPAR, not satisfy C1--C3 and C5) is created.
\subsubsection{Reward of the SA Subtask}
The internal reward $r_{t}^{\mathrm{u,int}}$ in the SA subtask is to encourage the agent to generate a suitable SA result (SSAR, satisfying C4 and C6). In the design of the SA action, constraint C6 is met. Only constraint C4 needs to be ensured. $r_{t}^{\mathrm{u,int}}$ is given by
\begin{align}
r_{t}^{\mathrm{u,int}}=\omega^{\mathrm{u,int}}\Gamma\left(\mathrm{C}4\right)
\end{align}
where $\omega^{\mathrm{u,int}}$ is the penalty coefficient. If constraint C${x}$ is satisfied, $\Gamma(\mathrm{C}x)=0$; otherwise, $\Gamma(\mathrm{C}x)=1$. For the joint reward $r_{t}^{\mathrm{u,jo}}$, the objective of $\emph{OP1}$ can be directly applied,
\begin{align}
r_{t}^{\mathrm{u,jo}}=\omega^{\mathrm{u,jo}}\exp\left(\omega^{\mathrm{jo}}\stackrel[m=1]{M}{\sum}\varpi_{m}\left(\stackrel[i=1]{N_{\mathrm{F}}}{\sum}\tilde{R}_{i,m}\right)\right)
\end{align}
where $\omega^{\mathrm{u,jo}}>0$ is the excitation coefficient, and $\omega^{\mathrm{jo}}>0$ is the adjustable factor. The final reward $r_{t}^{\mathrm{u}}$ of the SA-agent is $r_{t}^{\mathrm{u}}\!=\! r_{t}^{\mathrm{u,int}}\!+\!r_{t}^{\mathrm{u,jo}}$. The reward using an exponential function is in light of a ``reward shaping'' technique \cite{ng1999policy}.
\subsubsection{Reward of the PA Subtask}
The internal reward $r_{m,t}^{\mathrm{p,int}}$ of agent $m$ is to motivate the generation of SPAR (i.e., satisfying constraints C1--C3 and C5). The way in which the actual power value is calculated, i.e., (\ref{allocated power}), ensures that constraint C1 is satisfied, and the update algorithm of $\mathbf{v}_{m,t}$ ensures C3 is satisfied. Accordingly, $r_{m,t}^{\mathrm{p,int}}$ can be written as
\begin{align}
r_{m,t}^{\mathrm{p,int}}=\omega_{\mathrm{I}}^{\mathrm{p,int}}\Gamma\left(\mathrm{C}2\right)+\omega_{\mathrm{II}}^{\mathrm{p,int}}\left(\stackrel[i=1]{N_{\mathrm{F}}}{\sum}\tilde{R}_{i,m}-R_{m}^{\mathrm{min}}\right)
\end{align}
where $\omega_{\mathrm{I}}^{\mathrm{p,int}}\!<\!0$ is the penalty coefficient, and $\omega_{\mathrm{II}}^{\mathrm{p,int}}\!>\!0$ is an excitation coefficient.

In this paper, we decouple the reward in light of a ``difference reward'' technique \cite{tumer2007distributed}, where the joint reward $r_{m,t}^{\mathrm{p,jo}}$ of PA-agent $m$ is defined as
\begin{align}\label{weight_agent}
\resizebox{.85\hsize}{!}{$r_{m,t}^{\mathrm{p,jo}}\!=\!\left(\Theta_{m}\right)/\stackrel[m=1]{M}{\sum}\left(\Theta_{m}\right)\omega_{m}^{\mathrm{p,jo}}\exp\left(\omega_{m}^{\mathrm{jo}}\stackrel[m=1]{M}{\sum}\left(\varpi_{m}\stackrel[i=1]{N_{\mathrm{F}}}{\sum}\tilde{R}_{i,m}\right)\right),$}
\end{align}
where $\Theta_{m}\!=\!\varpi_{m}\sum_{m=1}^{N_{\mathrm{F}}}\tilde{R}_{i,m}$; and $\omega_{m}^{\mathrm{p,jo}}$ and $\omega_{m}^{\mathrm{jo}}$ adjust the magnitude of $r_{m,t}^{\mathrm{p,jo}}$. In (\ref{weight_agent}), the total optimization objective value is divided between different agents proportionally, depending the throughput and weight of each individual agent. In contrast, a total optimization objective value cannot be directly applied as the joint reward $r_{m,t}^{\mathrm{p,jo}}$ of each PA-agent for two reasons: 1) A global reward makes it difficult for each agent to deduce its individual contribution. The gradient computed for each actor does not explicitly reason about how the agent's actions contribute to the global reward \cite{foerster2017counterfactual}; and 2) different users (i.e., different agents) can have different weights to account for different priorities. The final reward $r_{m,t}^{\mathrm{p}}$ of PA-agent $m$ is $r_{m,t}^{\mathrm{p}}\!=\! r_{m,t}^{\mathrm{p,int}}\!+\!r_{m,t}^{\mathrm{p,jo}}$.

As discussed, having individual per-agent rewards can be better than having a single global reward. However, in the case where we do not have a-priori knowledge on the contribution of per-agent, existing algorithms, such as QMIX \cite{2018QMIX}, can be used to train the agents only with the global reward.
\subsection{Updating Algorithm Derivation of Neural Network in DRL}
The action-value function $Q(s_{t},a_{t})$ describes the expected return of an action $a_{t}$ taken in state $s_{t}$. Following the target policy $\pi$, $Q(s_{t},a_{t})$ is written as $Q^{\pi}(s_{t},a_{t})=\mathbb{E}_{r_{t},s_{t}\sim E,a_{t}\sim\pi}[\mathcal{R}_{t}|s_{t},a_{t}]$, where $E$ is the environment state distribution and $\pi$ is the target policy distribution. Policy $\pi$ may be either stochastic or deterministic. Let $\mu$ represent the deterministic target policy. Then, $\mu$ can be described as a function $\mu$: $\mathcal{S}\leftarrow\mathcal{A}$. By utilizing the recursive relationship (i.e., the Bellman equation), $Q(s_{t},a_{t})$ can be transformed into
\begin{align}\label{Q_update}
Q^{\mu}\left(s_{t},a_{t}\right)=\mathbb{E}_{r_{t},s_{t+1}\sim E}[r_{t}+\gamma Q^{\mu}\left(s_{t+1},\mu\left(s_{t+1}\right)\right)].
\end{align}

DQN is a popular type of DRL, and applies \emph{experience relay} and \emph{target network} techniques \cite{mnih2015human}. However, DQN only supports control problems with a relatively small set of low-dimensional and discrete actions \cite{lillicrap2015continuous}. By integrating the advantages of DQN and actor-critic (AC) architecture, DDPG can support continuous or high-dimensional action spaces \cite{silver2014deterministic}. The AC structure consists of an actor network (i.e., the policy network) and a critic network (i.e., the Q-network) \cite{konda2000actor}.

\begin{figure}[ht]
\centering{}\includegraphics[scale=0.4]{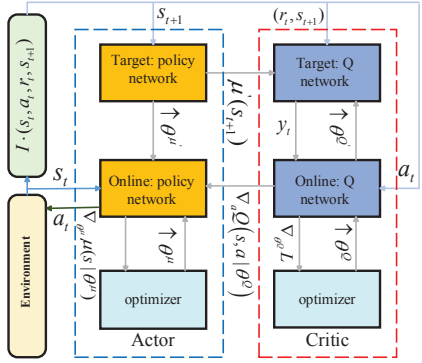}
\caption{DDPG diagram. $\uparrow$ indicates the update of the network parameters. The green box is the ``experience replay'' pool.}\label{DDPG}
\end{figure}

DDPG also adopts copy network, which creates a copy for each critic and actor network to improve stability and convergence. The original network is referred to as online network, and the copy network is referred to as target network. The target network is updated by using a ``soft update'' algorithm \cite{lillicrap2015continuous}. Fig. \ref{DDPG} shows the flow diagram of DDPG, where $I$ is the size of the ``experience replay'' pool.

Define $\theta^{Q}_{t}$ as the network parameters of the critic network. Based on the gradient update, the network can be trained by minimizing the loss functions $L_{t}(\theta_{t}^{Q})$. At each step $t$, $L_{t}(\theta_{t}^{Q})=
\mathbb{E}_{a_{t}\sim\pi,s_{t}\sim E}[(y_{t}\!-\!Q^{\mu}(s_{t},a_{t}|\theta_{t}^{Q}))^{2}]$, where $y_{t}\!=\!\mathbb{E}_{s_{t+1}\sim E}[r_{t}\!+\!\gamma\max_{a^{'}}Q^{\mu}(s_{t+1},a^{'})|\theta_{t-1}^{Q})]$. The gradient of the critic network is calculated by differentiating $L_{t}(\theta_{t}^{Q})$ with respect to $\theta^{Q}_{t}$ giving:
\begin{align}\label{single_Q}
\resizebox{.85\hsize}{!}{$\nabla_{\theta_{t}^{Q}}L_{t}(\theta_{t}^{Q})\!=\!\mathbb{E}_{a\sim\pi,s\sim E}[\nabla_{\theta_{t}^{Q}}Q^{\mu}(s_{t},a_{t})|\theta_{t}^{Q})(r_{t}\!+\!\gamma\max_{a^{'}}Q^{\mu}(s_{t+1},a^{'})|\theta_{t-1}^{Q})\!-\!Q^{\mu}(s_{t},a_{t})|\theta_{t}^{Q}))])$}.
\end{align}

The policy gradient of the actor network is calculated by employing the chain rule of differentiation to the expected future return from the initial distribution $J$ with respect to $\theta_{t}^{\mu}$ \cite{silver2014deterministic}, where $\theta_{t}^{\mu}$ is the actor parameters. This gives
\begin{align}\label{single_policy}
\nabla_{\theta_{t}^{\mu}}J&\approx\mathbb{E}_{s\sim E}[\nabla_{\theta_{t}^{\mu}}Q^{\mu}(s,a|\theta_{t}^{Q})|_{s=s_{t},a=\mu(s_{t}|\theta_{t}^{\mu})}]\\ \nonumber
&=\mathbb{E}_{s\sim E}[\nabla_{a}Q^{\mu}(s,a|\theta_{t}^{Q})|_{s=s_{t},a=\mu(s_{t})}\nabla_{\theta_{t}^{\mu}}\mu(s|\theta_{t}^{\mu})|_{s=s_{t}}]
\end{align}
where $J=\mathbb{E}_{r_{t},s_{t}\sim E,a_{t}\sim\pi}[\mathcal{R}_{1}]$. Degris \emph{et al.} \cite{degris2012off} have proved that this is a good approximation since it can preserve the set of local optima to which gradient ascent converges.
\section{Proposed DRL-JRM Framework}
\subsection{Global DRL-JRM Framework}
\begin{figure*}[htb]
\centering{}\includegraphics[scale=0.38]{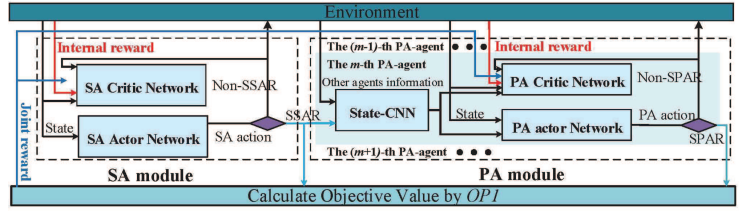}
\caption{The DRL-JRM framework.}\label{resource_stru}
\end{figure*}

The DRL-JRM framework consists of an \emph{SA module} and a \emph{PA module}, as shown in Fig.~\ref{resource_stru}.
The SA module adopts the single-agent technique and is responsible for the SA subtask. The input of the SA module includes the channel gains $\mathbf{H}$, user priorities $\mathbf{W}$, and QoS constraints $\mathbf{R}^{\mathrm{min}}$. The output is the SA result $\left\{ \mathbf{a}_{1}^{\mathrm{u}},\ldots,\mathbf{a}_{M}^{\mathrm{u}}\right\}$. The PA module employs the multi-agent technique and is responsible for the PA subtask. The input of the PA module includes the SA result $\left\{ \mathbf{a}_{1}^{\mathrm{u}},\ldots,\mathbf{a}_{M}^{\mathrm{u}}\right\}$, $\mathbf{H}$, $\mathbf{W}$ and $\mathbf{R}^{\mathrm{min}}$. The output is the PA result $\left\{ \mathbf{a}_{1}^{\mathrm{p}},\ldots,\mathbf{a}_{M}^{\mathrm{p}}\right\}$.

Given the ability of CNN in feature extraction and data compression, we design an state-CNN to extract  other agents information $\mathbf{s}_{\mathrm{other},m,t}^{\mathrm{p}}$ for information compression. The details of the state-CNN are provided in Section IV-C.

\textbf{Operation mechanism of DRL-JRM framework:} In each epoch (iteration), after $M$ steps, the SA actor network (SA-AN) outputs the SA action. If the SA action fails to meet constraints C4 or C6 (i.e., non-SSAR), a small internal reward is generated and will be input to the SA critic network (SA-CN) to update the action-value function $Q^{\mu}(s_{t},a_{t})$. According to $Q^{\mu}(s_{t},a_{t})$, SA-AN can be improved by the policy gradient method based on (\ref{single_policy}). The process repeats until the SA-agent can generate the SA action that meets constraints C4 and C6 (i.e., SSAR). Based on SSAR, after executing $T_{\mathrm{max}}^{\mathrm{PA}}$ steps, the PA results can be obtained by the $M$ PA-agents. Non-SPAR would lead to a small internal reward and, in turn, further learning of the PA-agents until the generation of SPAR. Note that SSAR and SPAR may not be optimal or convergent. We substitute the SSAR and SPAR into \emph{OP1} to calculate the objective value and obtain the joint reward, which are fed back to the SA-CN and PA critic network (PA-CN) to further improve the SSAR and SPAR, respectively. This repeats until the network converges with the JRM results.
\subsection{SA Module}
\begin{figure*}[htb]
\centering{}\includegraphics[scale=0.75]{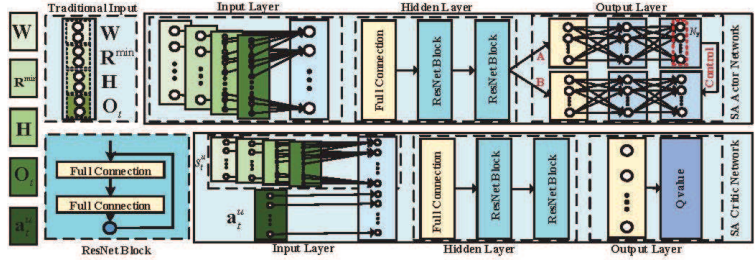}
\caption{The network detail structure of the SA actor network and the SA critic network.}\label{SA_stru}
\end{figure*}

Fig. \ref{SA_stru} provides the detailed network composition of the SA module. Three improvements are developed to improve efficiency and convergence, as will be shown in Section \uppercase\expandafter{\romannumeral5}-G.
\begin{Trick}
Input Design: \emph{Different from the traditional method of flattening all user information directly into a fully connected neural network, we design a hierarchical input method. In this method, the inputs of the neural network associated with an individual user are forwarded to a separate, localized, smaller neural network between the input and hidden layers of the overall neural network. All the localized, smaller neural networks output their results to a hidden layer to produce the results of the overall neural network.}
\end{Trick}
\begin{Trick}
ResNet: \emph{In general, we can obtain stronger network expression ability by increasing the number of layers. When the number of network layers is large, gradient dispersion becomes an inevitable problem. Since the residual network (ResNet) can effectively solve the problem, we leverage ResNet to replace the fully connected neural network (FCNN) to improve the number of neural network layers.}
\end{Trick}
\begin{Trick}
Output Design: \emph{The output layer is jointly implemented by two networks: ``A''  and ``B'' networks, which have the same network structure and input format. The ``A'' network is used to determine the number of subcarriers occupied by each user, and the ``B'' network generates a final (actual) SA result based on the output of the ``A'' network.}
\end{Trick}
\subsection{PA Module}
\begin{figure*}[htb]
	\centering{}\includegraphics[scale=0.55]{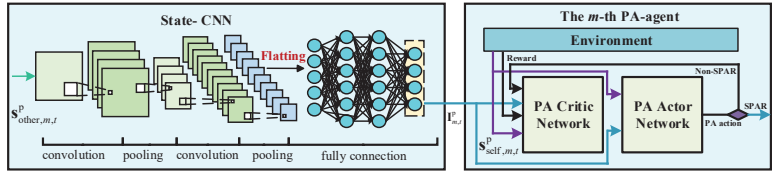}
	\caption{A pair of PA actor network and PA critic network in the PA module.}\label{PA_stru}
\end{figure*}

Fig. \ref{PA_stru} shows the  network composition of the $m$-th PA-agent in the PA module, where the detailed structures of PA-AN and PA-CN are similar to SA-AN and SA-CN.

In the PA module, we use a standard multi-agent learning paradigm: centralized training and decentralized execution \cite{kraemer2016multi}. A centralized action-value function method is applied, where the critic is augmented with extra information of other agents \cite{lowe2017multi}. Different from \cite{lowe2017multi}, our centralized action-value function of the $m$-th agent is given by
\begin{align}\label{Q_function_multi_agent_our}
&Q_{m}\left(\mathbf{s}_{\mathrm{self},m,t}^{\mathrm{p}},\mathbf{I}_{m,t}^{\mathrm{p}},\mathbf{a}_{1,t}^{\mathrm{p}},\cdots,\mathbf{a}_{M,t}^{\mathrm{p}}\right)\\ \nonumber
=&Q_{m}(\mathbf{s}_{\mathrm{self},m,t}^{\mathrm{p}},\mathbf{s}_{\mathrm{self},1,t}^{\mathrm{p}},\ldots,\mathbf{s}_{\mathrm{self},k,t}^{\mathrm{p}},\ldots\mathbf{s}_{\mathrm{self},M,t}^{\mathrm{p}},\mathbf{a}_{1,t}^{\mathrm{p}},\ldots,\mathbf{a}_{M,t}^{\mathrm{p}}),
\end{align}
where $k\neq m,$, by using the state-CNN, the observations of the other agents (i.e., $\mathbf{s}_{\mathrm{self},1,t}^{\mathrm{p}},\ldots,\mathbf{s}_{\mathrm{self},k,t}^{\mathrm{p}},\ldots,
\mathbf{s}_{\mathrm{self},M,t}^{\mathrm{p}}$) are compressed in $\mathbf{I}_{m,t}^{\mathrm{p}}$ which has the same dimension as the observation of the $m$-th agent $\mathbf{s}_{\mathrm{self},m,t}^{\mathrm{p}}$. By this means, we increase the proportion of the observation of the $m$-th agent in the observations of all agents. Specifically, the information of the other agents $\mathbf{s}_{\mathrm{other},m,t}^{\mathrm{p}}$ with size $(M-1)\times(3N_{\mathrm{F}}+2)$ is input into the state-CNN and converted to $\mathbf{I}_{m,t}^{\mathrm{p}}$ with size ($3N_{\mathrm{F}}+2$). So, $s_{m,t}^{\mathrm{p}}$ can be rewritten as $s_{m,t}^{\mathrm{p}}=\{ \mathbf{s}_{\mathrm{self},m,t}^{\mathrm{p}},(\mathbf{I}_{m,t}^{\mathrm{p}})^{\mathrm{T}}\} $.

According to \emph{centralized action-value function}, for PA actor network (PA-AN) with parameter $\theta_{m,t}^{\mathrm{p},\mu}$ of PA-agent $m$, policy gradient can be calculated by (\ref{PA policy grident}). The gradient of PA-CN with parameter $\theta_{m,t}^{\mathrm{p},Q}$ of PA-agent $m$ can be expressed by (\ref{PA loss}).
\begin{figure*}[htb]
\hrulefill
\begin{align}\label{PA policy grident}
\nabla_{\theta_{m,t}^{\mathrm{p},\mu}}J_{m}^{\mathrm{p}}&\approx\mathbb{E}_{s_{m}^{\mathrm{p}}\sim E}\left[\nabla_{\theta_{m,t}^{\mathrm{p},\mu}}Q_{m}^{\mathrm{p}}(s_{m}^{\mathrm{p}},\mathbf{a}_{1}^{\mathrm{p}},\ldots,\mathbf{a}_{M}^{\mathrm{p}}|\theta_{m,t}^{\mathrm{p},Q})|_{s_{m}^{\mathrm{p}}=s_{m,t}^{\mathrm{p}},\mathbf{a}_{m}^{\mathrm{p}}=\mu(s_{m,t}^{\mathrm{p}}|\theta_{m,t}^{\mathrm{p},\mu})}\right]\\ \nonumber
&=\mathbb{E}_{s_{m}^{\mathrm{p}}\sim E}[\nabla_{\mathbf{a}_{m,t}^{\mathrm{p}}}Q_{m}^{\mathrm{p}}(s_{m}^{\mathrm{p}},\mathbf{a}_{1}^{\mathrm{p}},\ldots,\mathbf{a}_{M}^{\mathrm{p}}|\theta_{m,t}^{\mathrm{p},Q})|_{s_{m}^{\mathrm{p}}=s_{m,t}^{\mathrm{p}},\mathbf{a}_{m}^{\mathrm{p}}=\mu(s_{m,t}^{\mathrm{p}})}\nabla_{\theta_{m,t}^{\mathrm{p},\mu}}\mu_{m}^{\mathrm{p}}(s_{m}^{\mathrm{p}}|\theta_{m,t}^{\mathrm{p},\mu})|_{s_{m}^{\mathrm{p}}=s_{m,t}^{\mathrm{p}}}].
\end{align}
\begin{align}\label{PA loss}
\nabla_{\theta_{m,t}^{\mathrm{p},Q}}L_{t}(\theta_{m,t}^{\mathrm{p},Q})&=\mathbb{E}_{\mathbf{a}_{m}^{\mathrm{p}}\sim\pi,s_{m}^{\mathrm{p}}\sim E}[(r_{m,t}^{\mathrm{p}}+\gamma\max_{\mathbf{a}_{m,t+1}^{\mathrm{p}}}Q_{m}^{\mathrm{p}}(s_{m,t+1}^{\mathrm{p}},\mu(s_{1,t+1}^{\mathrm{p}}),\ldots,\mu(s_{M,t+1}^{\mathrm{p}})|\theta_{m,t-1}^{\mathrm{p},Q})\\ \nonumber
&-Q_{m}^{\mathrm{p}}(s_{m,t}^{\mathrm{p}},\mu(s_{1,t}^{\mathrm{p}}),\ldots,\mu(s_{M,t}^{\mathrm{p}})|\theta_{m,t}^{\mathrm{p},Q}))\nabla_{\theta_{m,t}^{\mathrm{p},Q}}Q_{m}^{\mathrm{p}}(s_{m,t}^{\mathrm{p}},\mu(s_{1,t}^{\mathrm{p}}),\ldots,\mu(s_{M,t}^{\mathrm{p}})|\theta_{m,t}^{\mathrm{p},Q})].
\end{align}
\hrulefill
\end{figure*}

An agent may not have all observations of the others. We define an information perception degree (IPD) to measure the observations of the other PA-agents obtained by PA-agent $m$.
\begin{define}
The IPD of PA-agent $m$ is defined as
\begin{align}
\resizebox{.85\hsize}{!}{$\zeta_{m}=\left(\sum_{k=1}^{M}\sum_{l=1}^{N_{\mathrm{cate}}}I(\hat{x}_{k,l,m})\right)/\left(\sum_{k=1}^{M}\sum_{l=1}^{N_{\mathrm{cate}}}I(x_{k,l})\right),k\neq m,$}
\end{align}
where $N_{\mathrm{cate}}$ is the number of different types of observations and $N_{\mathrm{cate}}=5$ (i.e., user priority, QoS constraint, channel gain, SA result and the PA result of the current decision step). $I(x_{k,l})$ is the number of observations of the $l$-th ($\ensuremath{l}\in\{1,\cdots,N_{\mathrm{cate}}\}$) type for user $k$, i.e., the number of elements in matrix $\mathbf{s}_{\mathrm{other},m,t}^{\mathrm{p}}$. For example, the number of the observations on the channel gain (i.e., the third type of observation) is $N_{\mathrm{F}}$ for user $k$, then $I(x_{k,3})\!=\!N_{\mathrm{F}}$. $I(\hat{x}_{k,l,m})$ indicates that the PA-agent $m$ can obtain the number of observations of the $l$-th type for user $k$, that is, the number of non-zero entries in $\mathbf{s}_{\mathrm{other},m,t}^{\mathrm{p}}$. If the observation cannot be obtained, the corresponding entry in $\mathbf{s}_{\mathrm{other},m,t}^{\mathrm{p}}$ is 0; otherwise, the entry is 1.
\end{define}
\subsection{Model Training}
The training is summarized in Algorithms 1 and 2. $EM^{\mathrm{u}}$ and $EM_{m}^{\mathrm{p}}$ are the experience pools for the SA- and PA-agents.

As describe in Fig. \ref{resource_stru}, the DRL-JRM algorithm consists of five modules: SA-CN, SA-AN, state-CNN, PA-CN and PA-AN. We first define the following hyper-parameters of the network: $N_{\mathrm{full}}$ is the number of neurons in the fully connected layer; $d_{\mathrm{Res}}$ is the number of the ResNet blocks; $d_{\mathrm{cnn}}$ is the number of flattened layers in state-CNN; $D$ is the number of all convolutional layers; $l$ is the $l$-th convolutional layer; $M_{l}$ is the side length of the output feature map of convolution kernel in the $l$-th convolution layer; $K_{l}$ is the side length of each convolution kernel in the $l$-th convolution layer; and $C_{l}$ is the number of output channels of the $l$-th convolutional layer. The time- and space-complexities are provided in Theorem 2.
\begin{theo}
The time-complexity of DRL-JRM training is
\begin{align}
\resizebox{.85\hsize}{!}{$O(N_{\mathrm{ep}}N_{\mathrm{max}}^{\mathrm{SA}}N_{\mathrm{full}}M\left(4d_{\mathrm{Res}}N_{\mathrm{full}}+N_{\mathrm{full}}+4N_{\mathrm{F}}\right) $}
\nonumber  \\
\resizebox{.85\hsize}{!}{$+N_{\mathrm{ep}}N_{\mathrm{max}}^{\mathrm{PA}}T_{\mathrm{max}}^{\mathrm{PA}}(\stackrel[l=1]{D}{\sum}\left(M_{l}^{2}K_{l}^{2}C_{l-1}C_{l}\right)+\left(d_{\mathrm{net}}N_{\mathrm{full}}+14N_{\mathrm{F}}\right)N_{\mathrm{full}})),
,$}
\end{align}
The space-complexity of DRL-JRM training algorithm is
\begin{align}
\resizebox{.85\hsize}{!}{$O(N_{\mathrm{full}}\left(4d_{\mathrm{Res}}N_{\mathrm{full}}+N_{\mathrm{full}}+4N_{\mathrm{F}}\right)+18M\cdot N_{\mathrm{F}}\cdot EM^{\mathrm{u}}$}
\nonumber  \\
\resizebox{.85\hsize}{!}{$+\stackrel[l=1]{D}{\sum}\left(K_{l}^{2}C_{l-1}C_{l}\right)+\stackrel[l=1]{D}{\sum}\left(M^{2}C_{l}\right)+\left(d_{\mathrm{net}}N_{\mathrm{full}}+14N_{\mathrm{F}}\right)N_{\mathrm{full}})
,$}
\end{align}
where $d_{\mathrm{net}}=4d_{\mathrm{Res}}+1+d_{\mathrm{cnn}}$.
\end{theo}
\begin{IEEEproof}
Please refer to Appendix A.
\end{IEEEproof}
\section{Experimental Results}
For comparison purposes, the following benchmarks are tested: 1) DUCPA \cite{7557079}, which first groups users into clusters and then optimizes their respective powers, where SA is not considered and we use the greedy search method is used for SA for fair comparison; 2) CFJPBA \cite{celik2017cluster}, which first transforms cluster formation into a multi-partite matching problem and then solves the joint power-bandwidth allocation into a convex form by geometric programming. To make CFJPMA comparable with the proposed DRL-JRM, we first perform SA by greedy search, and then set the bandwidth to be evenly divided; 3) Random user grouping and power allocation (RUSPA), which first performs random SA and then applies the optimal PA algorithm proposed by Yang \emph{et al.} \cite{yang2017optimality}; 4) JPCA \cite{7417761}, which utilizes the Lagrangian duality to relax the individual power constraint and constructs the subproblem of Lagrangian relaxation. Then, a two-phase dynamic programming-based approach is applied; 5) JPSA \cite{7842087}, which runs an SCA-based algorithm to generate feasible solutions; 6) JPCA-DRL \cite{8790780}, which alternately solves the PA problem given a channel assignment, and runs the DRL to assign channels under the given PA results; and 7) JUSPA \cite{8755843}, which solves the PA problem with fractional quadratic transformation, and schedules users with a heuristic method in an alternating manner until convergence. When the optimization problem is small-scale, we can also employ the Gurobi solver supporting the B\&B method to obtain the optimal solutions.

\begin{algorithm}[htb]
\small
\caption{\small{Training the DRL-JRM.}}
\label{alg:Framwork}
\begin{algorithmic}[1] %show the number in each line
\STATE Initialize network parameters
\FOR {Epoch = 1,2,...,$N_{\mathrm{ep}}$}
\STATE Initialize a random process for the exploration of SA action and PA action
\STATE Receive initial state $s_{t}^{\mathrm{u}}$ and $s_{m,t}^{\mathrm{p}}$ for all $m$
\STATE Execute \textbf{SA-AN} in SA module with a total of $M$ steps
\STATE Get a complete SA policy $a_{t}^{\mathrm{u}}$
\IF {$a_{t}^{\mathrm{u}}$ is suitable SA policy}
\STATE Execute \textbf{PA-AN} in PA module
\STATE Get a complete PA policy $\mathbf{a}_{m,t}^{\mathrm{p}}$
\IF {$\mathbf{a}_{m,t}^{\mathrm{p}}$ is suitable PA policy}
\STATE Calculate the \emph{OP1}, and produce joint reward $r_{t}^{\mathrm{u,jo}}$ and $r_{m,t}^{\mathrm{p,jo}}$
\STATE Calculate total reward $r_{t}^{\mathrm{u}}$ and $r_{m,t}^{\mathrm{p}}$
\STATE Observe the next SA state $s_{t+1}^{\mathrm{u}}$ and PA state $s_{m,t+1}^{\mathrm{p}}$
\STATE Store $\left(s_{t}^{\mathrm{u}},a_{t}^{\mathrm{u}},r_{t}^{\mathrm{u}},s_{t+1}^{\mathrm{u}}\right)$ to $EM^{\mathrm{u}}$, and store $(s_{m,t}^{\mathrm{p}},\mathbf{a}_{m,t}^{\mathrm{p}},r_{m,t}^{\mathrm{p}},s_{m,t\:+\:1}^{\mathrm{p}})$ to  the $m$-th experience replay pool $EM_{m}^{\mathrm{p}}$
\IF {$EM^{\mathrm{u}}$ is full}
\STATE \textbf{Procedure 1}: \emph{\textbf{Training SA network}}
\STATE \textbf{Procedure 2}: \emph{\textbf{Training PA network}}
\ENDIF
\ELSE
\STATE Run \textbf{PA-agents} in Algorithm 2
\ENDIF
\ELSE
\STATE Run \textbf{SA-agent} in Algorithm 2
\ENDIF
\ENDFOR
\STATE
\STATE \textbf{Procedure 1}: \emph{\textbf{Training SA network}}
\STATE Sample a random mini-batch of transitions from $EM^{\mathrm{u}}$
\STATE Update SA-CN and SA-AN based on update algorithms (e.g., (\ref{single_Q}) and (\ref{single_policy}))
\STATE ``Soft update'' the target networks
\end{algorithmic}
\end{algorithm}
\begin{algorithm}[htb]
\small
\caption{\small{Network update algorithms.}}
\label{alg:Framwork}
\begin{algorithmic}[1] %show the number in each line
\STATE Run \textbf{SA-agent:}
\FOR {Episode = 1,2,...,$N_{\mathrm{max}}^{\mathrm{SA}}$}
\FOR {$t$ = 1,2,...,$M$}
\STATE Input $s_{t}^{\mathrm{u}}$ to SA-AN and output $a_{t}^{u}$.
\STATE Acquire internal reward $r_{t}^{\mathrm{u,int}}$ and  calculate total reward $r_{t+1}^{\mathrm{u}}$
\STATE Observe the next SA state $s_{t+1}^{\mathrm{u}}$
\STATE Store $\left(s_{t}^{\mathrm{u}},a_{t}^{u},r_{t}^{\mathrm{u}},s_{t+1}^{\mathrm{u}}\right)$ to $EM^{\mathrm{u}}$
\ENDFOR
\IF {$EM^{\mathrm{u}}$ is full}
\STATE \textbf{Procedure 1}: \emph{\textbf{Training SA network}}
\ENDIF
\ENDFOR
\STATE
\STATE Run \textbf{PA-agents:}
\FOR {Episode = 1,2,...,$N_{\mathrm{max}}^{\mathrm{PA}}$}
\FOR {$t$ = 1,2,...,$T_{\mathrm{max}}^{\mathrm{PA}}$}
\STATE For each agent $m$, input $s_{m,t}^{\mathrm{p}}$ to RAN and output $\mathbf{a}_{m,t}^{\mathrm{p}}$
\STATE Acquire internal reward $r_{m,t}^{\mathrm{p,int}}$ and calculate total reward $r_{m,t+1}^{\mathrm{p}}$
\STATE Observe the next PA state $s_{m,t+1}^{\mathrm{p}}$
\STATE Store $(s_{m,t}^{\mathrm{p}},\mathbf{a}_{m,t}^{\mathrm{p}},r_{m,t}^{\mathrm{p}},s_{m,t\:+\:1}^{\mathrm{p}})$ to  the $m$-th experience replay pool $EM_{m}^{\mathrm{p}}$.
\IF {$EM_{m}^{\mathrm{p}}$ is full}
\STATE \textbf{Procedure 2}: \emph{\textbf{Training PA network}}
\ENDIF
\ENDFOR
\ENDFOR
\STATE
\STATE \textbf{Procedure 2}: \emph{\textbf{Training PA network}}
\FOR {agent $m$ = 1,...,$M$}
\STATE Sample a random mini-batch of transitions from $EM_{m}^{\mathrm{p}}$
\STATE Update PA-AN and PA-CN of PA-agent $m$ by the update algorithms (e.g., (\ref{PA policy grident}) and (\ref{PA loss}))
\STATE ``Soft update'' the target networks
\ENDFOR
\end{algorithmic}
\end{algorithm}
\subsection{Simulation Setup}
Taking the BS location as the center, the users are distributed inside a concentric circle with the radius ranging from 30 m to 300 m, and follow the spatial Poisson point process $\varPhi(\lambda)$ with density $\lambda\!=\!2$. 3GPP urban path loss model is adopted with the path loss exponent of 3.6. The noise power is $\sigma_{m}^{2}\!=\!\frac{BN_{0}}{N_{\mathrm{F}}}$, where $B\!=\!5$ MHz, $N_{\mathrm{F}}\!=\!64$, and the noise power spectral density $N_{0}\!=\!-173$ dBm. The different requirements $R_{m}^{\mathrm{min}}\!\sim\!\mathcal{N}\left(80\:\mathrm{kbits}/\mathrm{s},10\right)$. IPD $\zeta_{m}\!=\!1.00, \forall m$, unless otherwise specified. Both the SA-AN and SA-CN contain three ResNet blocks. As shown in Fig. \ref{SA_stru}, each ResNet block is composed of two FCNN layers, where each FCNN layer has 128 neurons, and the corresponding activation function is ReLU (i.e., $\max\left(0,x\right)$). The output layer of the SA-AN is as described in Section \uppercase\expandafter{\romannumeral4}-B, and the last layer has only a single neuron and has no activation function. The penultimate layer has 64 neurons, and the activation function is also ReLU. Similarly, all PA-ANs and PA-CNs of the PA-agents have the same hidden and output layers structures as the SA-AN and SA-CN. For the state-CNN, the hidden layers are convolution (ReLU)+max pooling, convolution (ReLU)+max pooling, fully connection (ReLU), or fully connection (ReLU). The output layer of the state-CNN has $3N_{\mathrm{F}}+2$ neurons, and the activation function is the Sigmoid function (i.e., $1/\left(1+e^{-x}\right)$). The \emph{RMSProp} algorithm is used to conduct gradient descent. Both $EM^{\mathrm{u}}$ and $EM_{m}^{\mathrm{p}}$ are set to 5,000 and 4,000. The batch size is 128. The learning rates of SA-AN and SA-CN are 0.001 and 0.003. The learning rate of PA-AN and PA-CN of PA-agent $m$ are 0.002 and 0.005. Other hyper-parameters are set as follows: $\omega^{\mathrm{u,int}}\!=\!-5$, $\omega_{\mathrm{I}}^{\mathrm{p,int}}\!=\!-8$, $\omega_{\mathrm{II}}^{\mathrm{p,int}}\!=\!3$, $\omega^{\mathrm{u,jo}}\!=\!1.5$, $\omega^{\mathrm{jo}}\!=\!0.25$, $\omega_{\mathrm{m}}^{\mathrm{p,jo}}\!=\!16$, and $\omega_{\mathrm{m}}^{\mathrm{jo}}\!=\!0.45$. $N_{\mathrm{max}}^{\mathrm{SA}}\!=\!20,000$, $N_{\mathrm{max}}^{\mathrm{PA}}\!=\!35,000$, $N_{\mathrm{ep}}\!=\!15,000$, $T_{\mathrm{max}}^{\mathrm{PA}}\!=\!100$, $\gamma^{\mathrm{u}}\!=\!\gamma_{m}^{\mathrm{p}}\!=\!0.99$, and $\vartheta=10^{-5}$. These hyper-parameters are set up based on extensive simulation tests and actual accuracy requirements.
\begin{table*}[htp]
\small
	\centering{}
	\textbf{Table \uppercase\expandafter{\romannumeral2}}~~A comprehensive comparison between the proposed algorithm and existing studies. The results of AT are provided for those with the same objective as and similar settings to the proposed algorithms, and not applicable (n.a.) for the rest of the existing studies.\\
\scriptsize
\setlength{\tabcolsep}{0.850mm}{
		\begin{tabular}{|c|c|c|c|c|}
        \hline
        \multirow{1}*{List} & \multirow{1}*{\textbf{Assumption}} & \multirow{1}*{\textbf{Objective}} & \multirow{1}*{\textbf{Methodology}}& \multirow{1}*{\textbf{AT(bit/s/Hz)}}\\
        \hline
        \tabincell{c}{This \\paper} & \tabincell{c}{imperfect SIC, multi-carrier, DL, multiple users per\\ cluster, hardware sensitivity requirement considered} & \tabincell{c}{power allocation, user pair-\\ing, subcarrier scheduling}& \tabincell{c}{deep reinforcement learning}&8.9419\\
			\hline
        \cite{7587811}, 2016  & \multirow{7}{*}{\tabincell{c}{perfect SIC,\\ multi-carrier,\\ DL,\\ multiple users per cluster,\\ hardware sensitivity requirement not considered}} & \tabincell{c}{power and channel allocation}& \tabincell{c}{Lagrangian duality, \\dynamic programming}& n.a.\\
		\cline{1-1} \cline{3-5}
        \cite{8403100}, 2018  & ~ & \tabincell{c}{power and subcarrier all-\\ocation, user clustering, } & \tabincell{c}{Lagrangian dual, sequenti- \\al convex programming}&n.a.\\
		\cline{1-1} \cline{3-5}
        \cite{7417761}, 2015  & ~ & \tabincell{c}{power and channel allocation}& \tabincell{c}{Lagrangian duality,\\ dynamic programming}&8.6588\\
		\cline{1-1} \cline{3-5}
        \cite{7557079}, 2016  & ~ & \tabincell{c}{power allocation,\\ user clustering} & \tabincell{c}{Karush-Kuhn-Tucker\\ optimality conditions}&7.8928\\
			\hline
        \cite{7934461}, 2017  & \tabincell{c}{imperfect SIC, multi-carrier, DL, two users per \\cluster, hardware sensitivity requirement not considered} & \tabincell{c}{power and rate allocation, \\ user scheduling}& \tabincell{c}{B\&B approach, difference\\ of convex programming}&n.a.\\
			\hline
        \cite{8540884}, 2019 & \multirow{5}*{\tabincell{c}{imperfect SIC, multi-carrier,\\ DL, multiple users per cluster,\\ hardware sensitivity requirement not considered}}  & \tabincell{c}{power allocation}& \tabincell{c}{fractional quadratic\\ transformation}&n.a.\\
		\cline{1-1} \cline{3-5}
        \cite{8755843}, 2019 & ~  & \tabincell{c}{power allocation, \\ user scheduling}& \tabincell{c}{fractional quadratic trans-\\formation, heuristic}&8.6145\\
		\cline{1-1} \cline{3-5}
        \cite{celik2017cluster}, 2017  & ~ & \tabincell{c}{power allocation,\\ user clustering} & \tabincell{c}{multi-partite matching,\\ geometric programming}&7.5795\\
			\hline
        \cite{8790780}, 2019  &  \tabincell{c}{perfect SIC, multi-carrier, DL, two users per\\ cluster, hardware sensitivity requirement not considered} & \tabincell{c}{power allocation, \\user pairing} & \tabincell{c}{water-filling algorithm, \\deep reinforcement learning}&8.7405\\
			\hline
        \cite{7842087}, 2016  & \tabincell{c}{perfect SIC, multi-carrier, DL, two users per\\ cluster, hardware sensitivity requirement not considered}  & \tabincell{c}{power allocation\\ subcarrier allocation}& \tabincell{c}{monotonic optimization}&8.7648\\
        \hline
	\end{tabular}}
\end{table*}

Table II compares the proposed approach with  the existing works which are the most relevant. The average throughput (AT) results are simulated and compared between the proposed approach and those works, where the number of users is 44, the total transmit power is 42 dBm, the hardware sensitivity requirement is 0.08 dBm, and the SIC error factor is $10^{-4}$. More detailed comparisons are provided in the following.
\begin{figure*}[htb]
  \centering
  \subfigure[]{%\label{fig:subfig1:a} %% label for first subfigure
    \includegraphics[width=2.3in]{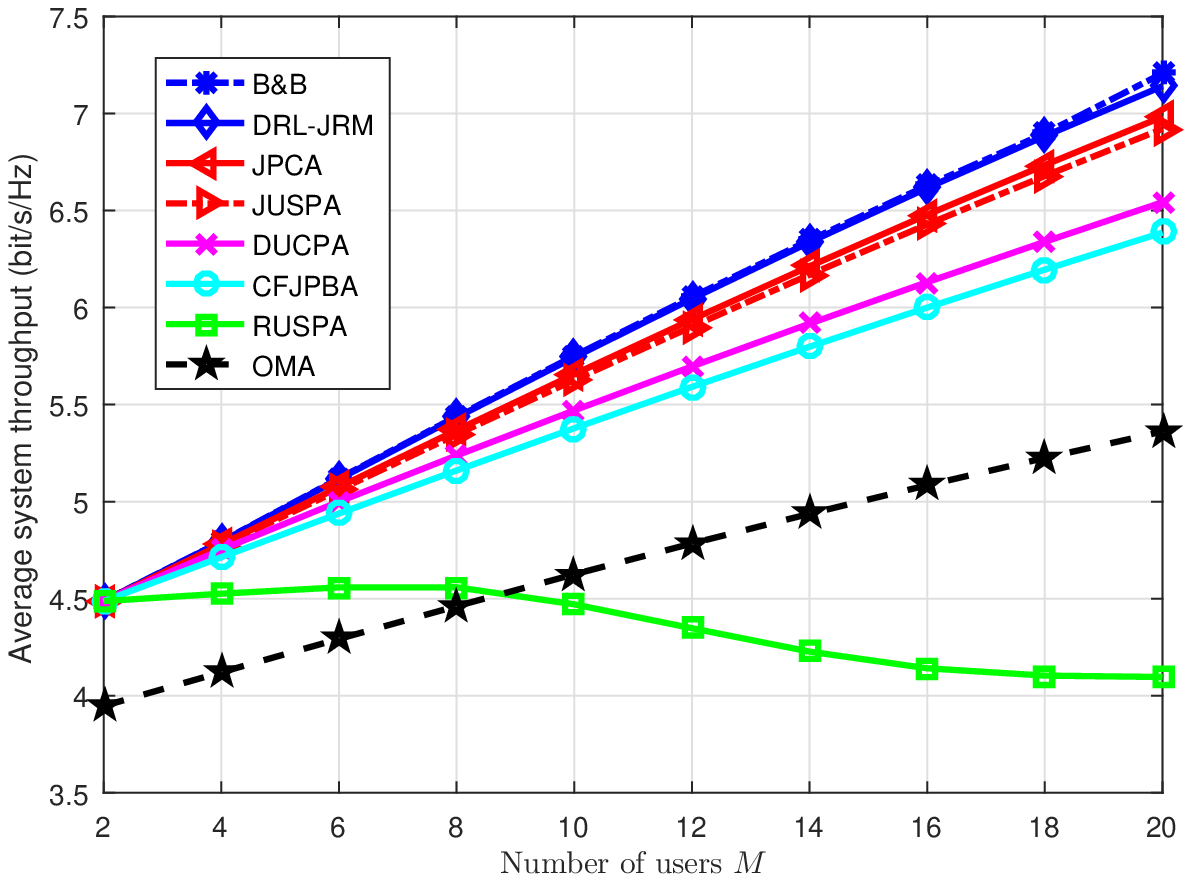}
  }
  \subfigure[]{%\label{fig:subfig1:b} %% label for second subfigure
    \includegraphics[width=2.3in]{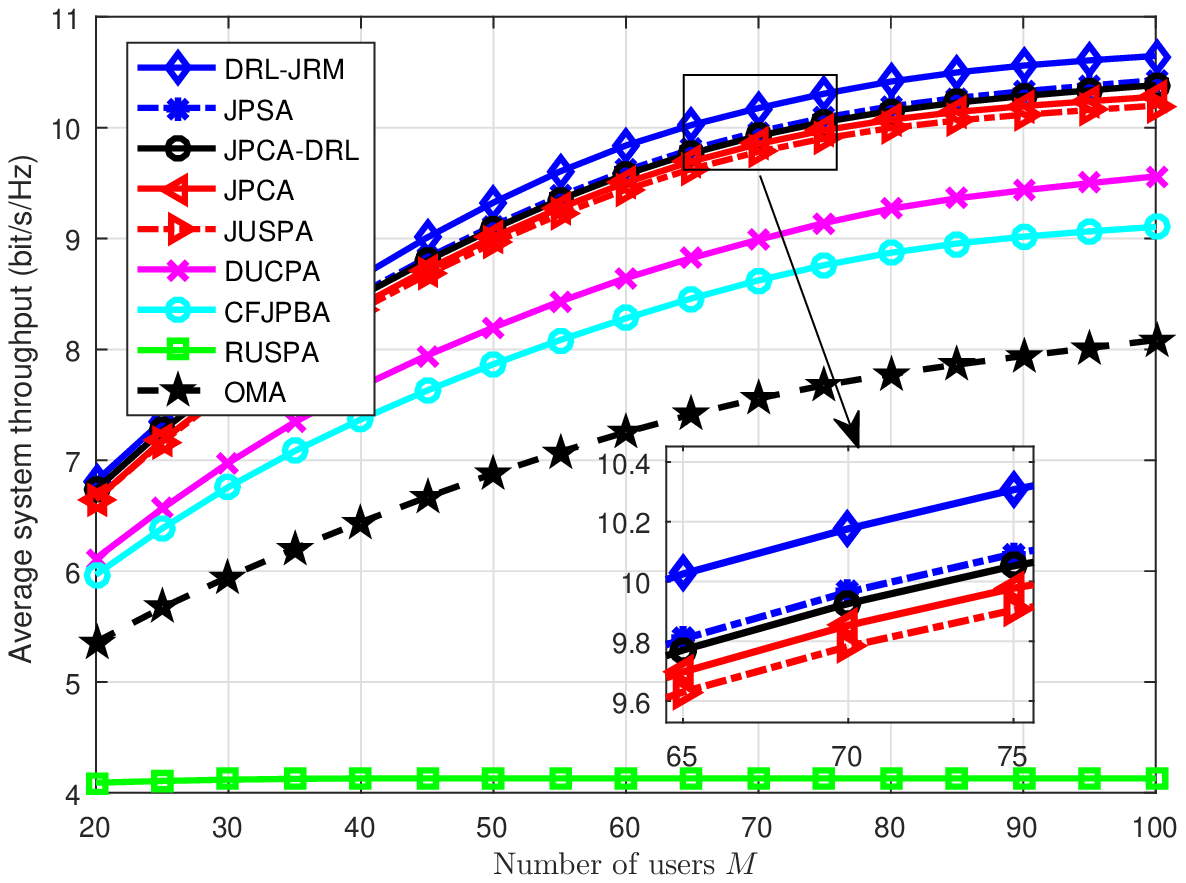}
  }
  \subfigure[]{%\label{fig:subfig1:b} %% label for second subfigure
    \includegraphics[width=2.3in]{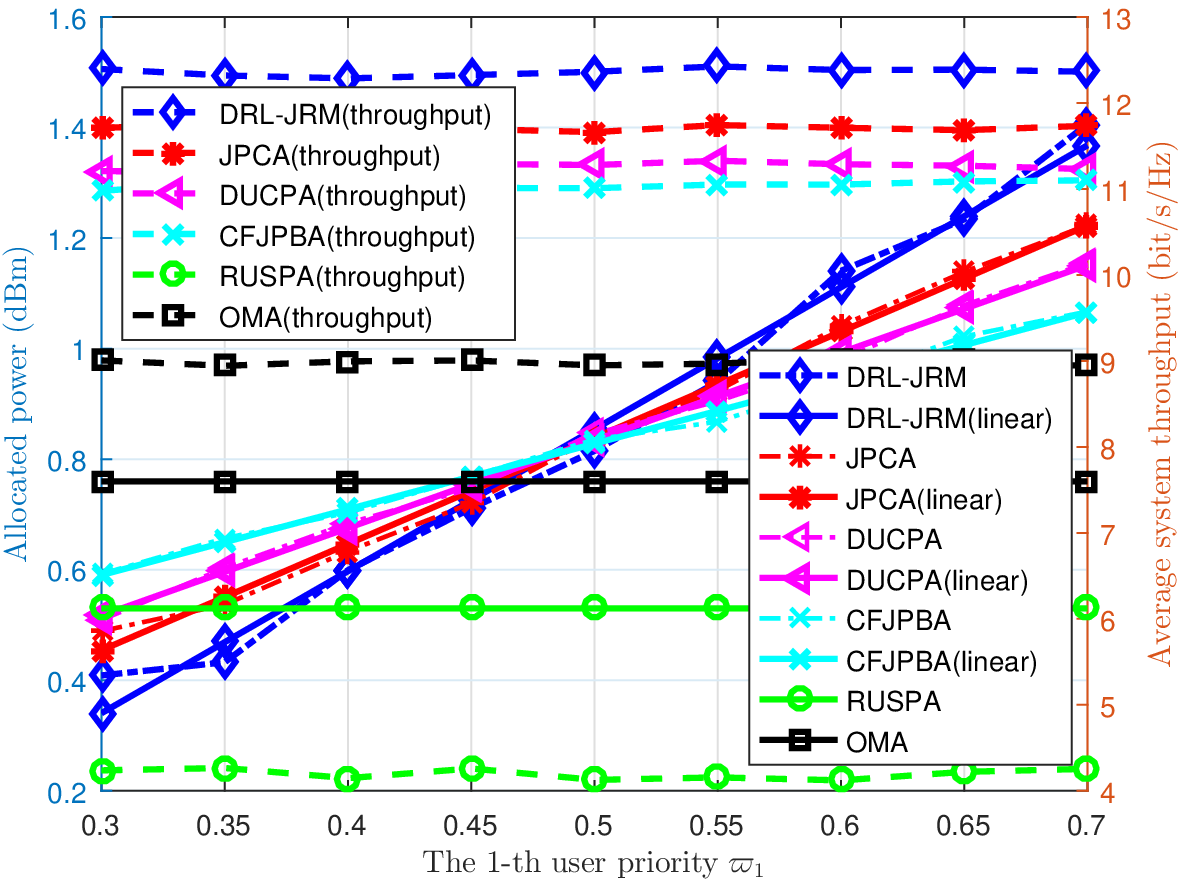}
  }
  \caption{(a)--(b): Comparison of AT versus user number. (a) $N_{\mathrm{F}}\!=\!8$, $P_{\mathrm{total}}\!=\!40$ dBm, $N_{\mathrm{max}}\!=\!4$, $p_{\Delta}\!=\!0.05$ dBm and $\varepsilon^{2}\!=\!10^{-4}$. (b) $N_{\mathrm{F}}\!=\!64$, $P_{\mathrm{total}}\!=\!42$ dBm, $N_{\mathrm{max}}\!=\!2$, $p_{\Delta}\!=\!0.08$ dBm and $\varepsilon^{2}\!=\!10^{-4}$. (c): $N_{\mathrm{F}}\!=\!20$, $M\!=\!60$, $N_{\mathrm{max}}=4$, $P_{\mathrm{total}}\!=\!46$ dBm, $p_{\Delta}\!=\!0.05$ dBm and $\varepsilon^{2}\!=\!10^{-3}$. $\varpi_{1}\!=\!0.1$ is selected as the reference time. we linearly fit the allocated power values, and the fitting results are marked with the label ``linear''.}
  \label{A_throughput_user_number} %% label for entire figure
\end{figure*}

\begin{figure*}[t]
  \centering
  \subfigure[]{%\label{fig:subfig1:a} %% label for first subfigure
    \includegraphics[width=2.3in]{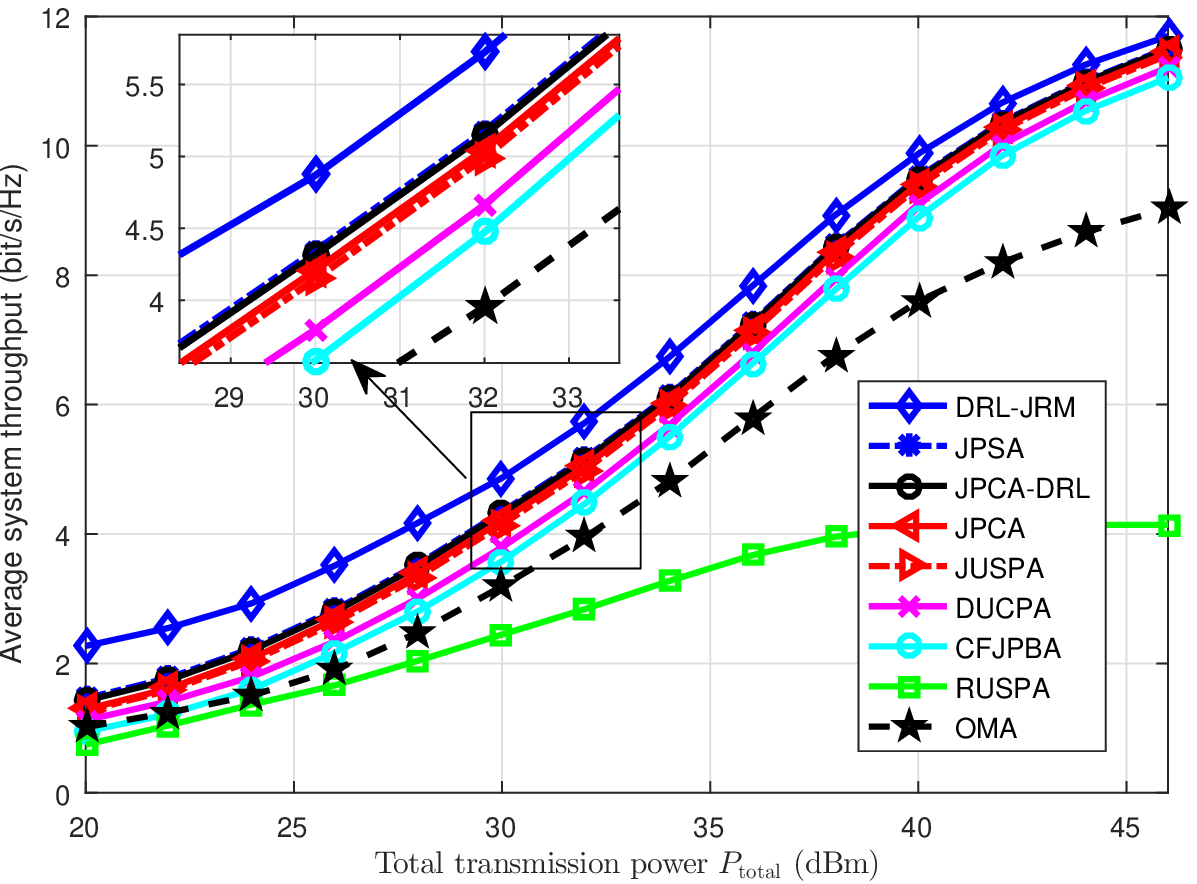}
  }
  \subfigure[]{%\label{fig:subfig1:b} %% label for second subfigure
    \includegraphics[width=2.3in]{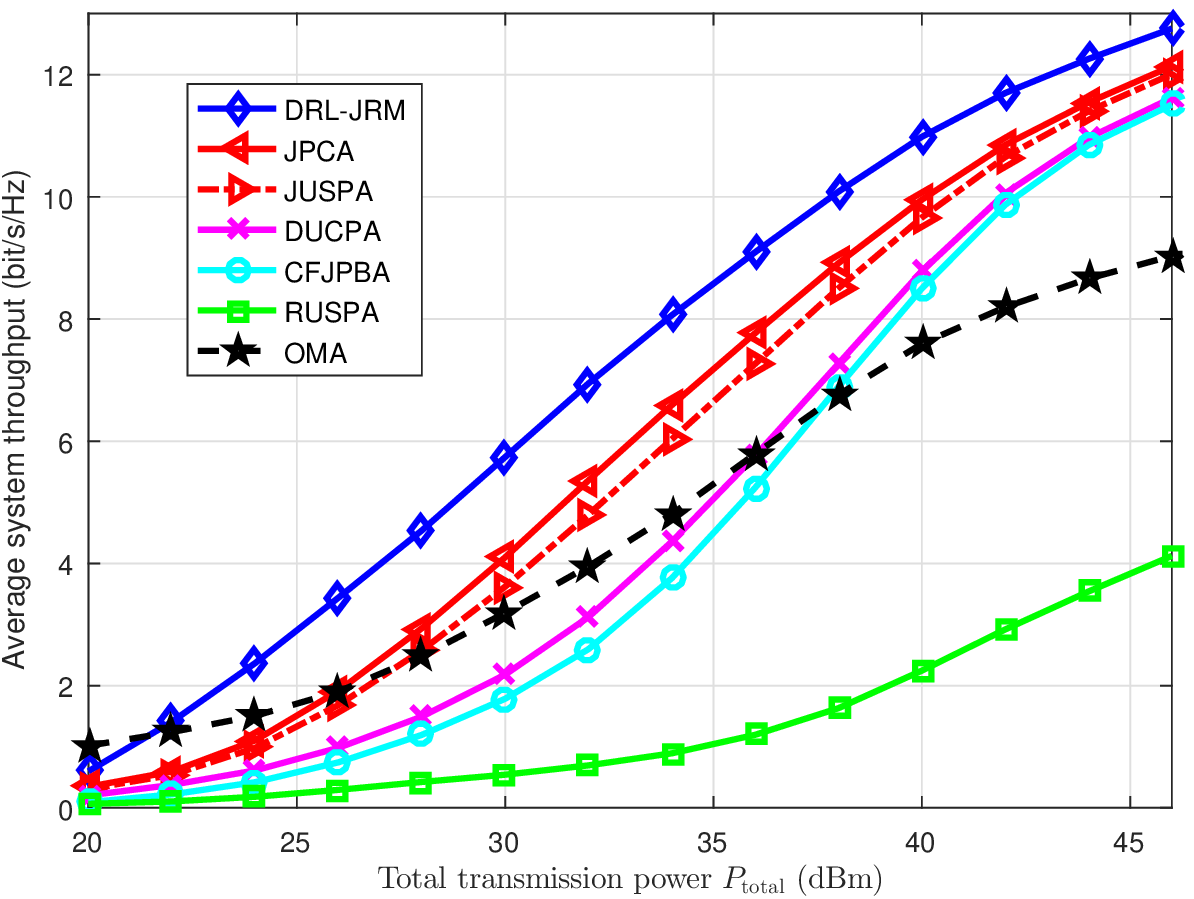}
  }
   \subfigure[]{%\label{fig:subfig1:a} %% label for first subfigure
    \includegraphics[width=2.3in]{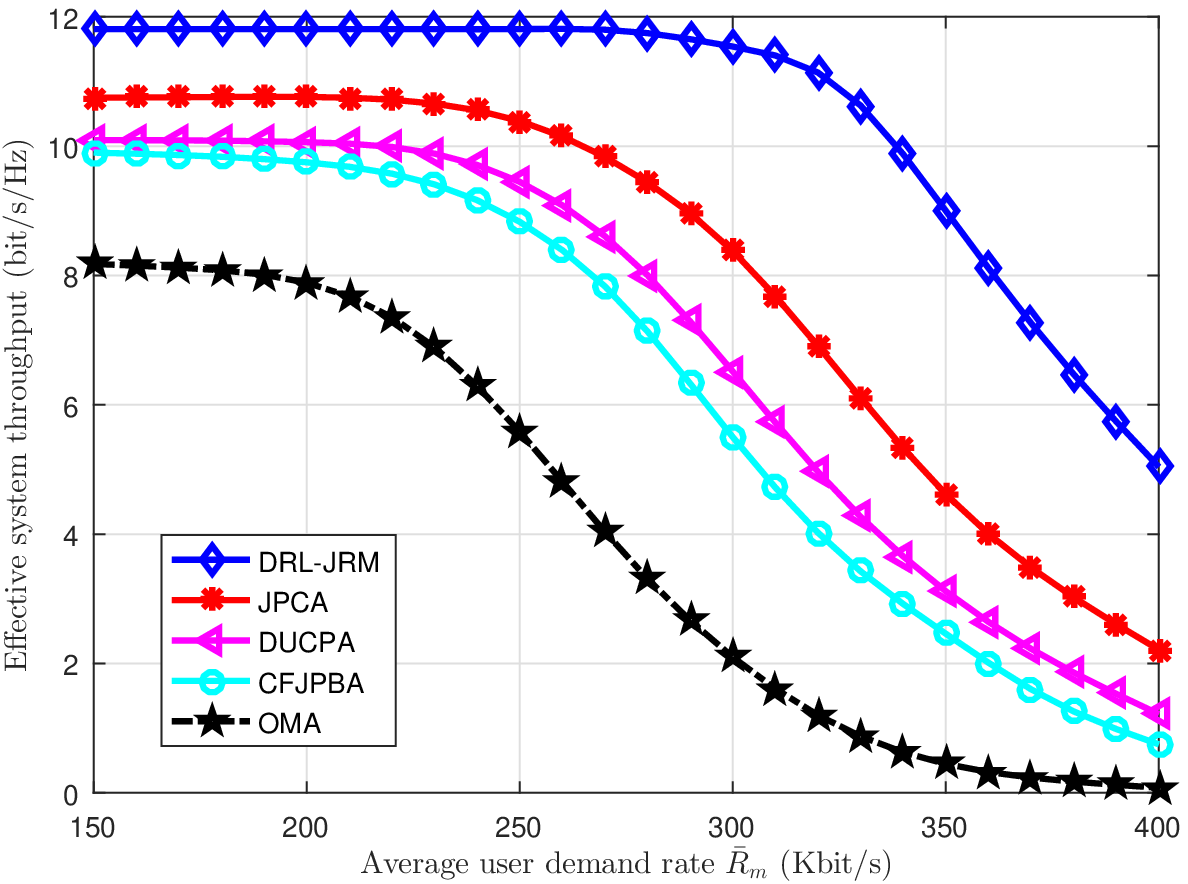}
  }
  \subfigure[]{%\label{fig:subfig1:b} %% label for second subfigure
    \includegraphics[width=2.3in]{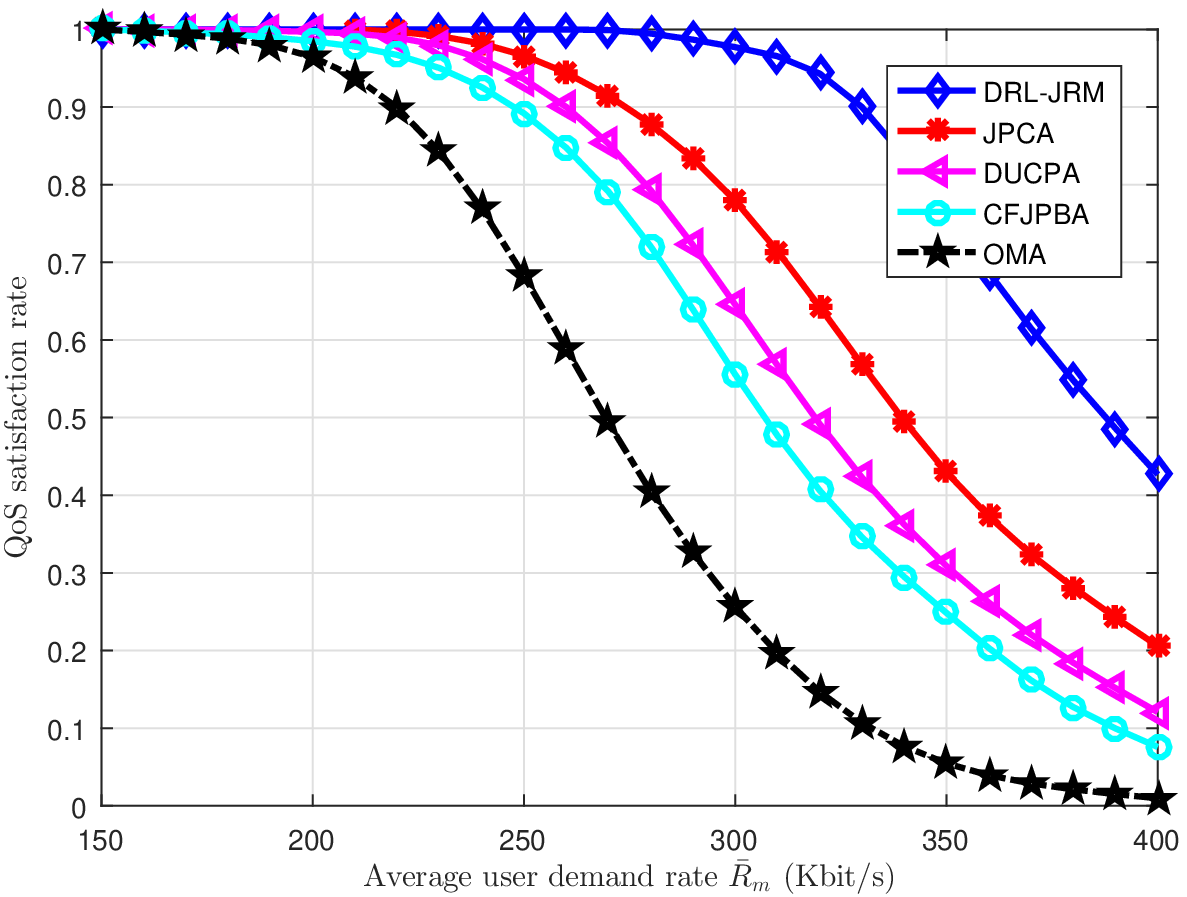}
  }
    \subfigure[]{%\label{fig:subfig1:a} %% label for first subfigure
    \includegraphics[width=2.3in]{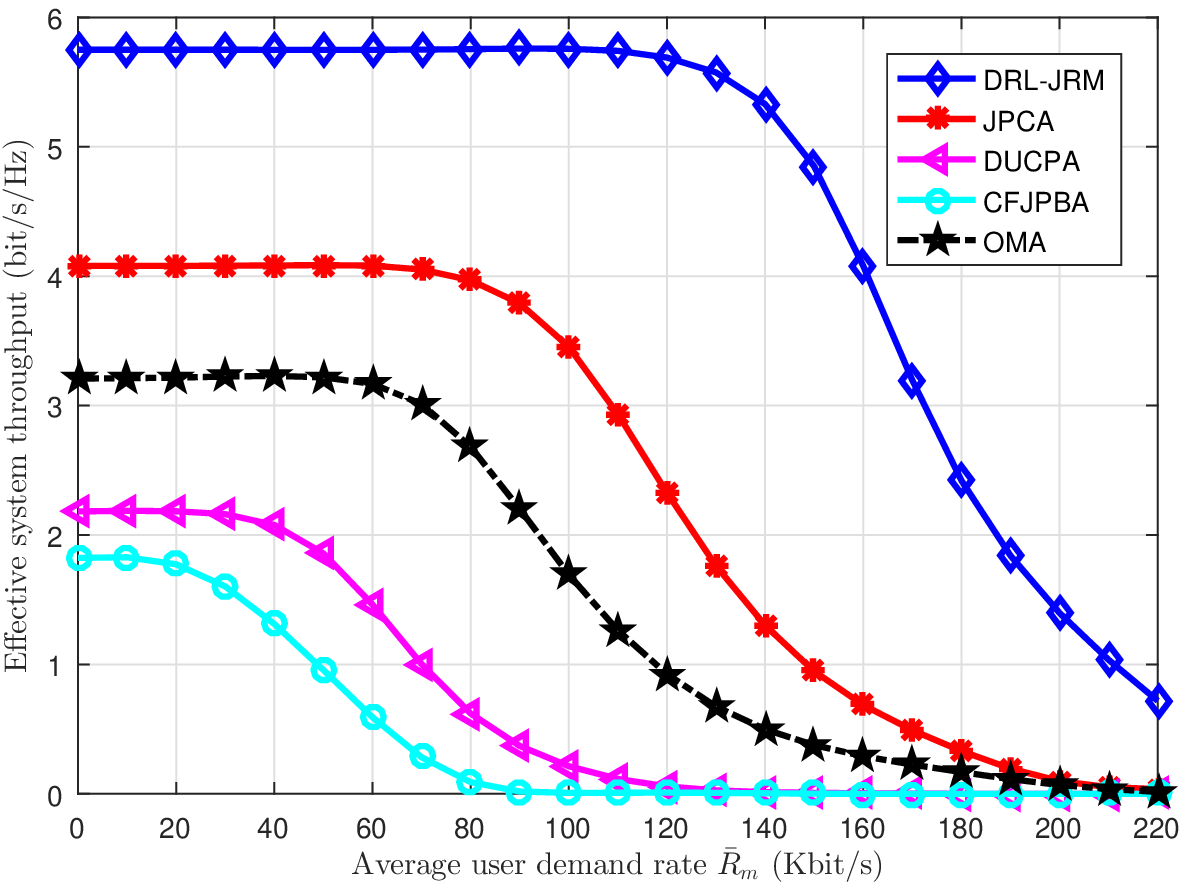}
  }
  \subfigure[]{%\label{fig:subfig1:b} %% label for second subfigure
    \includegraphics[width=2.3in]{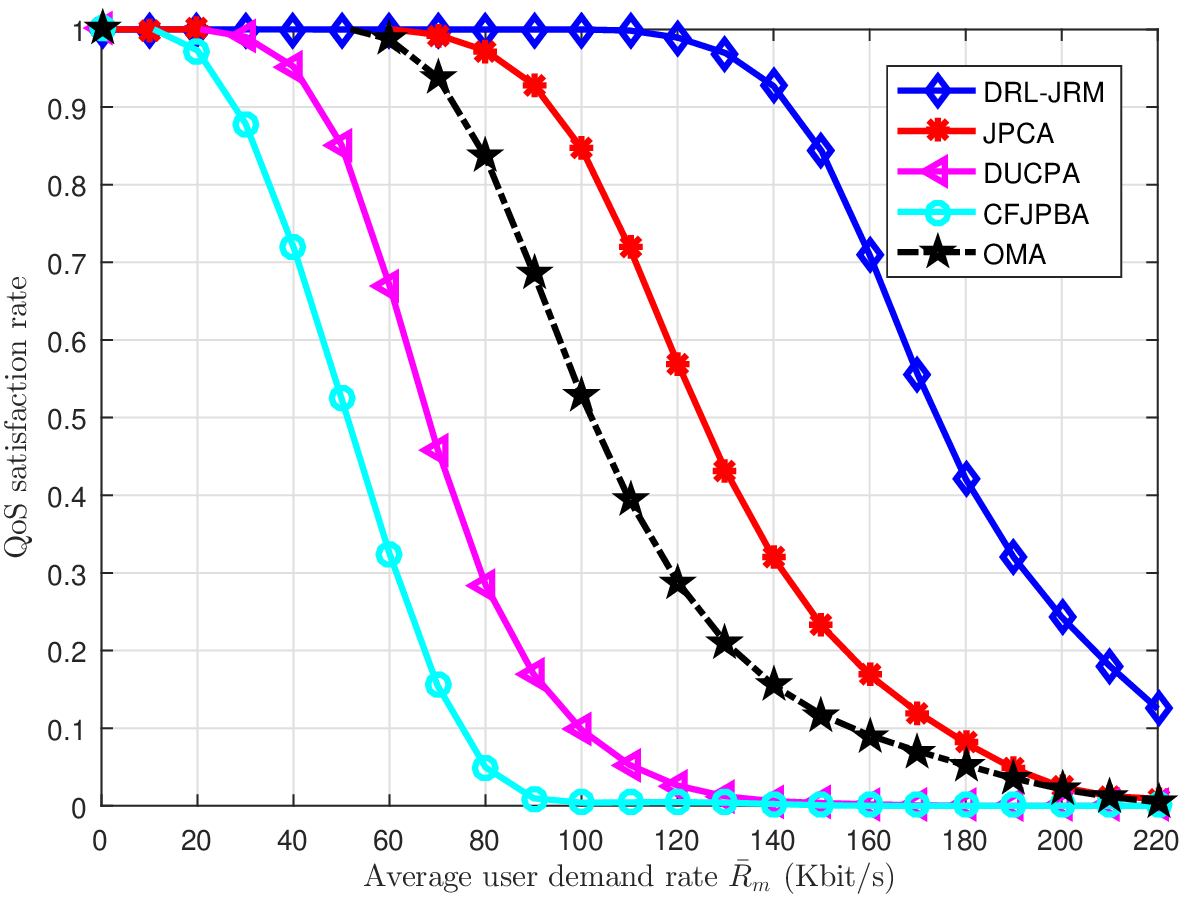}
  }
  \caption{(a)--(b):Comparison of AT versus total transmit power under different methods. (a) $N_{\mathrm{F}}=40$, $M=60$, $N_{\mathrm{max}}=2$, $p_{\Delta}=0.05$ dBm and $\varepsilon^{2}=10^{-2}$. (b) $N_{\mathrm{F}}=20$, $M=60$, $N_{\mathrm{max}}=4$, $p_{\Delta}=0.08$ dBm and $\varepsilon^{2}=10^{-2}$. (c)--(f):Comparison of $Q_{\mathrm{eff}}$ and $\varrho_{\mathrm{QoS}}$ versus user demand rate $R_{m}^{\mathrm{min}}$ under different methods. The abscissa is the average of demand rates (i.e., $\bar{R}_{m}\!=\!\frac{1}{M}\sum_{m=1}^{M}R_{m}^{\mathrm{min}}$). (c) with (d) $N_{\mathrm{F}}\!=\!20$, $M\!=\!60$, $N_{\mathrm{max}}\!=\!4$, $P_{\mathrm{total}}\!=\!42$ dBm, $p_{\Delta}\!=\!0.05$ dBm and $\varepsilon^{2}\!=\!10^{-3}$. (e) with (f) $N_{\mathrm{F}}\!=\!20$, $M\!=\!60$, $N_{\mathrm{max}}\!=\!4$, $P_{\mathrm{total}}\!=\!30$ dBm, $p_{\Delta}\!=\!0.08$ dBm and $\varepsilon^{2}\!=\!10^{-2}$.}
  \label{A_throughput_power} %% label for entire figure
\end{figure*}
\subsection{Average System Throughput Versus Total Number of Users}
Fig. \ref{A_throughput_user_number}(a) compares the AT of the considered algorithms in a small-scale problem ($2\leq M\leq20$). The proposed DRL-JRM method is better than the existing alternatives and very close to the optimal result obtained by the B\&B method. Fig. \ref{A_throughput_user_number}(b) compares the AT under a much larger problem setting, and DRL-JRM performs the best in all methods. Since $N_{\mathrm{max}}\!=\!2$ in Fig. \ref{A_throughput_user_number}(b), JPSA and JPCA-DRL serve as the baseline.

In Fig. \ref{A_throughput_user_number}(b), the proposed DRL-JRM method is better than JPSA and JPCA-DRL in terms of AT. This is because despite the CSI is imperfect at the base station, JPCA-DRL treats it as the perfect CSI for user selection, subcarrier and power allocation (as done in \cite{8790780}). On the other hand, the AT of the resulting user selection and resource allocation is evaluated under the actual CSI. Moreover, the number of users multiplexed per subcarrier is the same across subcarriers in JUSPA \cite{8755843}, reducing flexibility. OMA achieves a lower throughput due to its less efficient use of the spectrum.
\subsection{Sensitivity to User Priority}
$\varpi_{m}$ accounts for the priority of different personalized requests.  Fig. \ref{A_throughput_user_number}(c) shows the changes of the allocated power for the first user (i.e., $m=1$) and the AT under different $\varpi_{1}$, where the priorities of the other users are randomly generated and remain unchanged. As $\varpi_{1}$ increases, the powers allocated to the first user by DRL-JRM, JPCA, DUCPA and CFJPBA increase significantly. The slope of DRL-JRM is the largest, which means that DRL-JRM is more sensitive to the change of $\varpi_{m}$ and has stronger personalized service ability.
\subsection{Average System Throughput Vs. Maximum Transmit Power}
Figs. \ref{A_throughput_power}(a) and (b) plot the AT versus  $P_{\mathrm{total}}$ when $N_{\mathrm{max}}\!=\!2$ and 4. When $P_{\mathrm{total}}$ is small, DRL-JRM exhibits a considerable gap over the other techniques. The gap decreases with the increase of $P_{\mathrm{total}}$, and the impact of co-subcarrier interference and PDSC diminishes. This reveals the benefit of DRL-JRM in resource allocation under imperfect SIC and PDSC, especially in the presence of strong co-subcarrier interference. In Fig. \ref{A_throughput_power}(b),  MC-NOMA schemes perform worse than the MC-OMA schemes when $P_{\mathrm{total}}=$20 dBm. This is because of: 1) the strong co-subcarrier interference ($N_{\mathrm{max}}\!=\!4$) and SIC error ($\varepsilon^{2}\!=\!10^{-2}$); and 2) the limit of PDSC (i.e., higher power is required to ensure the basic requirements of SIC). The AT of DRL-JRM is higher than the other algorithms under different $P_{\mathrm{total}}$. Not only does this confirm the adaptability of DRL-JRM to PDSC and imperfect SIC, but also indicates that a higher $P_{\mathrm{total}}$ is required when $N_{\mathrm{max}}$ is large. From Figs. \ref{A_throughput_user_number}(a), \ref{A_throughput_user_number}(b), \ref{A_throughput_power}(a) and \ref{A_throughput_power}(b), despite RUSPA utilizes the optimal PA, AT is low. This indicates the importance of a proper SA.
\subsection{Effective System Throughput Versus User Demand Rate}
\begin{figure*}[htb]
  \centering
  \subfigure[]{%\label{fig:subfig1:a} %% label for first subfigure
    \includegraphics[width=1.6in]{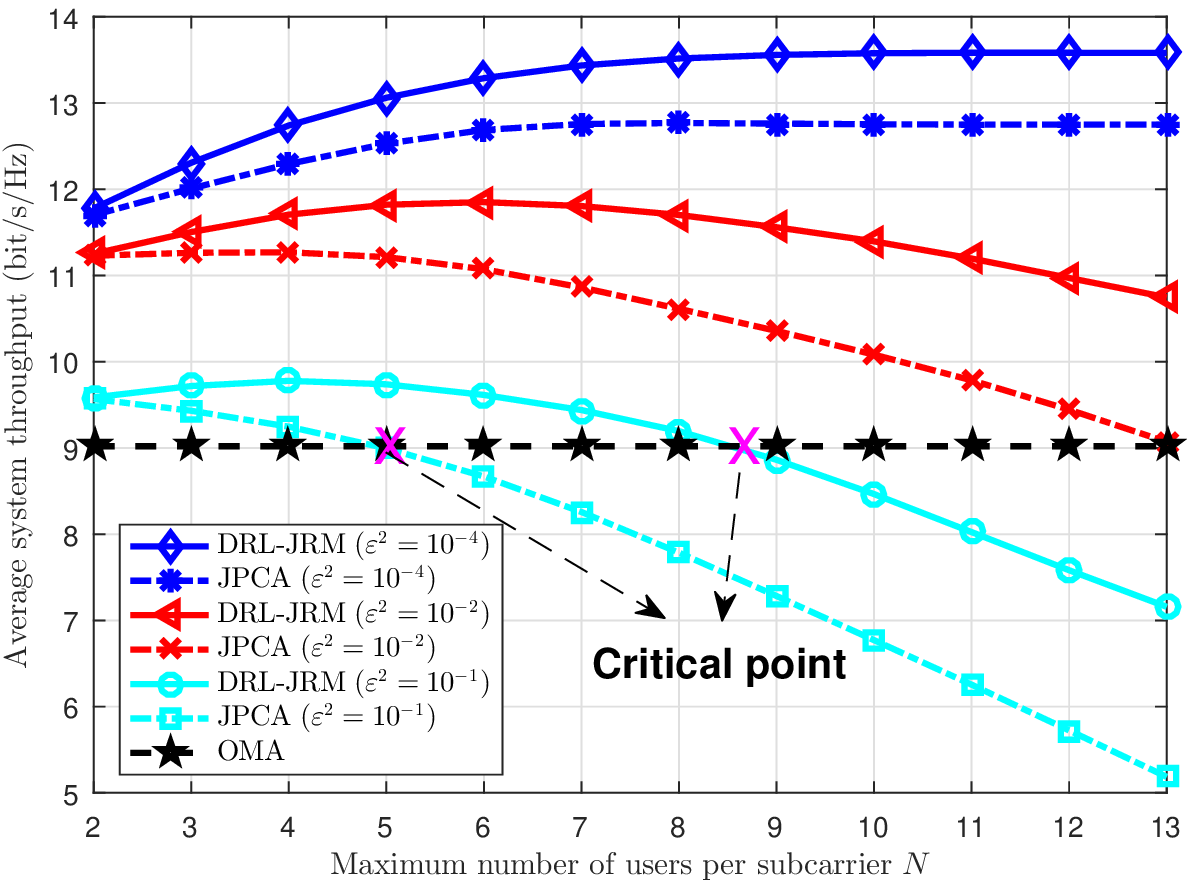}
  }
  \subfigure[]{%\label{fig:subfig1:b} %% label for second subfigure
    \includegraphics[width=1.6in]{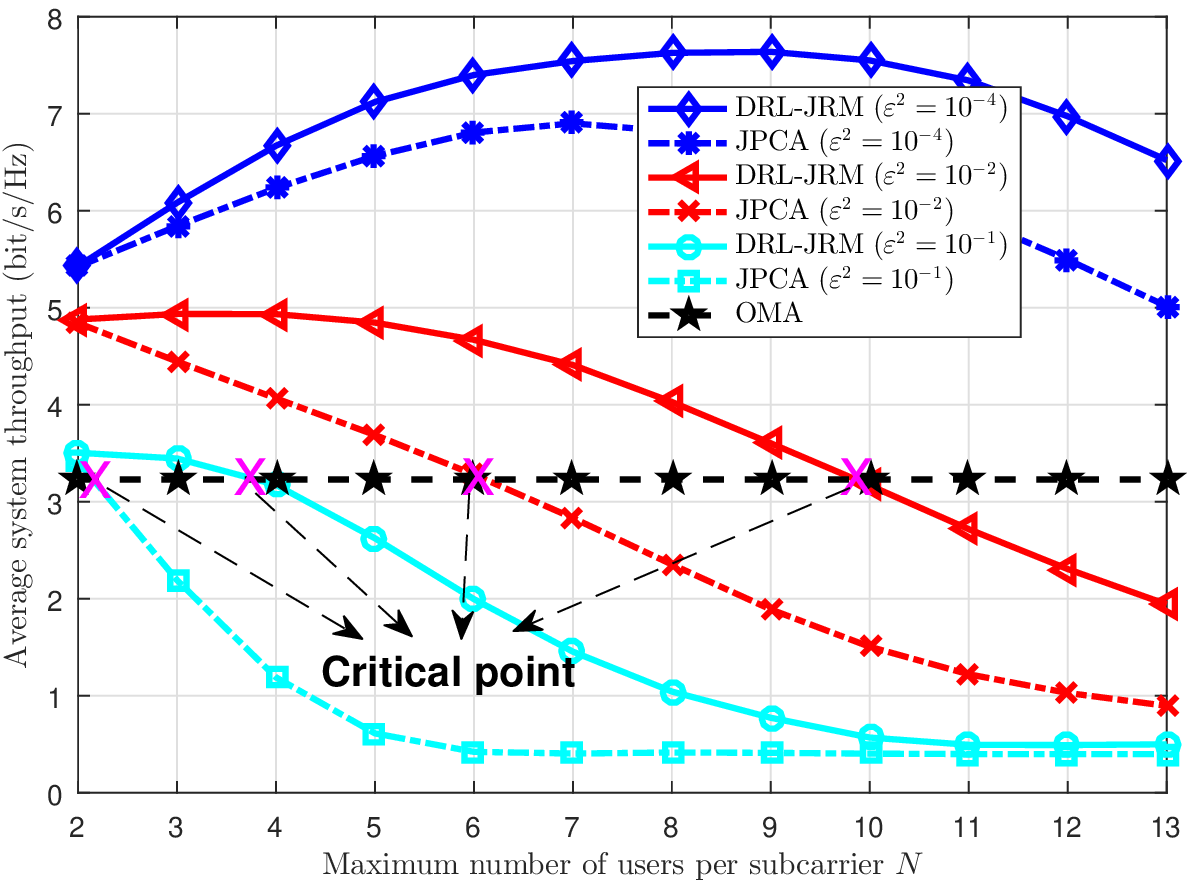}
  }
    \subfigure[]{%\label{fig:subfig1:a} %% label for first subfigure
    \includegraphics[width=1.6in]{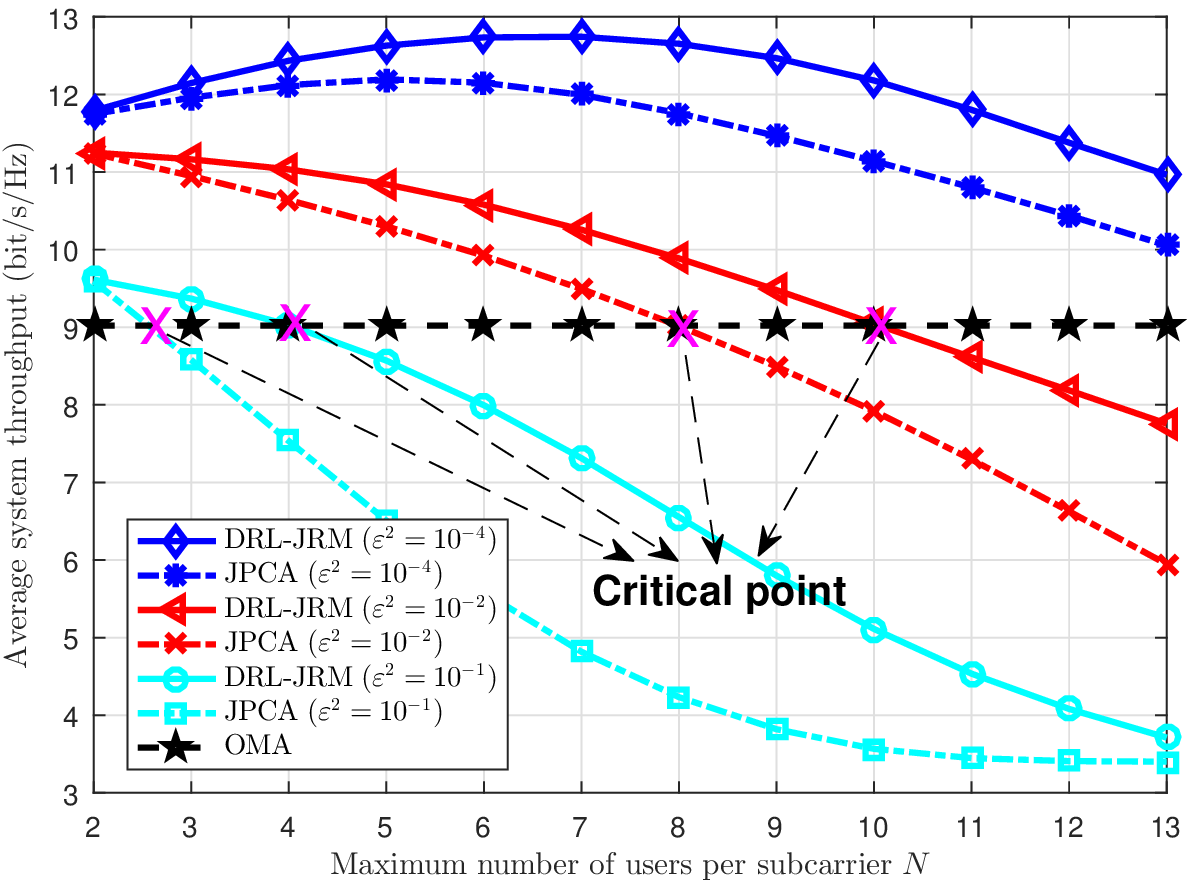}
  }
  \subfigure[]{%\label{fig:subfig1:b} %% label for second subfigure
    \includegraphics[width=1.6in]{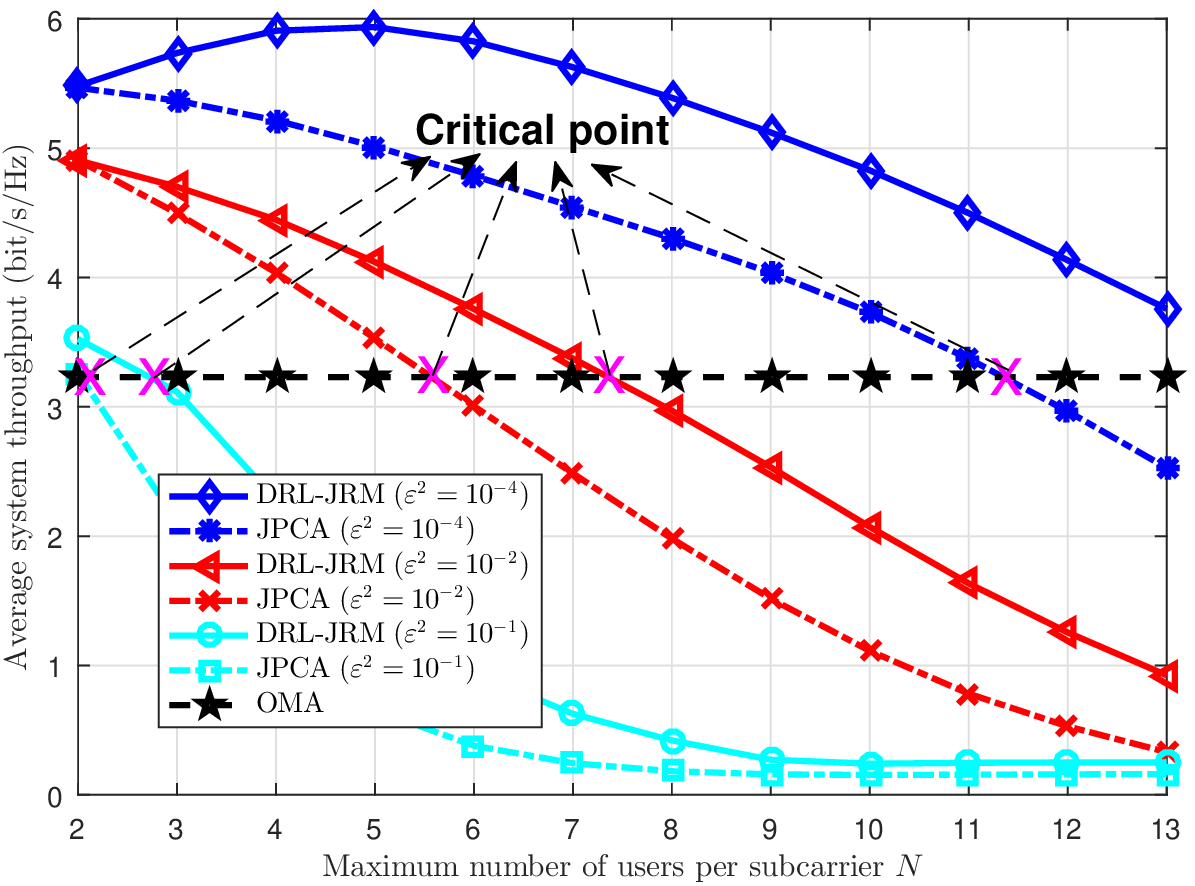}
  }
  \caption{AT versus $N_{\mathrm{max}}$ under different interference conditions. (a) $N_{\mathrm{F}}\!=\!40$, $M\!=\!60$, $P_{\mathrm{total}}\!=\!46$ dBm, and $p_{\Delta}\!=\!0.05$ dBm; (b) $N_{\mathrm{F}}\!=\!40$, $M\!=\!60$, $P_{\mathrm{total}}\!=\!30$ dBm, and $p_{\Delta}\!=\!0.05$ dBm; (c) $N_{\mathrm{F}}\!=\!40$, $M\!=\!60$, $P_{\mathrm{total}}\!=\!46$ dBm, and $p_{\Delta}\!=\!0.15$ dBm; (d) $N_{\mathrm{F}}\!=\!40$, $M\!=\!60$, $P_{\mathrm{total}}\!=\!30$ dBm, and $p_{\Delta}\!=\!0.15$ dBm.}
  \label{N_change} %% label for entire figure
\end{figure*}
The QoS satisfaction of users is also an important indicator. We define \emph{effective system throughput} $Q_{\mathrm{eff}}$ to measure the system throughput that meets the QoS requirements:
\begin{align}
\resizebox{.85\hsize}{!}{$Q_{\mathrm{eff}}=\stackrel[m=1]{M}{\sum}\left(\frac{1}{2}\stackrel[i=1]{N_{\mathrm{F}}}{\sum}\tilde{R}_{i,m}(\mathrm{sgn}(\stackrel[i=1]{N_{\mathrm{F}}}{\sum}\tilde{R}_{i,m}-R_{m}^{\mathrm{min}})+1)\right)
$}
\end{align}
where $\mathrm{sgn}\left(\cdot\right)$ is an \emph{extended} signum function. If $x\geq0$, $\mathrm{sgn}\left(x\right)=1$; otherwise, $\mathrm{sgn}\left(x\right)=-1$. In addition, the \emph{QoS satisfaction rate} $\varrho_{\mathrm{QoS}}$, is defined as
\begin{align}
\varrho_{\mathrm{QoS}}=\stackrel[m=1]{M}{\sum}\left(\frac{1}{2}(\mathrm{sgn}(\stackrel[i=1]{N_{\mathrm{F}}}{\sum}\tilde{R}_{i,m}-R_{m}^{\mathrm{min}})+1)\right)/M.
\end{align}
Fig. \ref{A_throughput_power}(c)--(f) compares the average effective system throughput and $\varrho_{\mathrm{QoS}}$ with the growth of $R_{m}^{\mathrm{min}}$. In the presence of weak interference, as shown in Figs. \ref{A_throughput_power}(c) and (d), DRL-JRM declines the slowest. DRL-JRM can effectively balance user QoS constraints while maintaining high system throughput, and hence achieve adequate resource allocation. In the presence of strong interference, as shown in Figs. \ref{A_throughput_power}(e) and (f), DRL-JRM declines the slowest, indicating its efficient use of resources.
 \begin{figure*}[htb]
  \centering % the size of figure is 7*25
  \subfigure[]{%\label{fig:subfig1:a} %% label for first subfigure
    \includegraphics[width=4.5in]{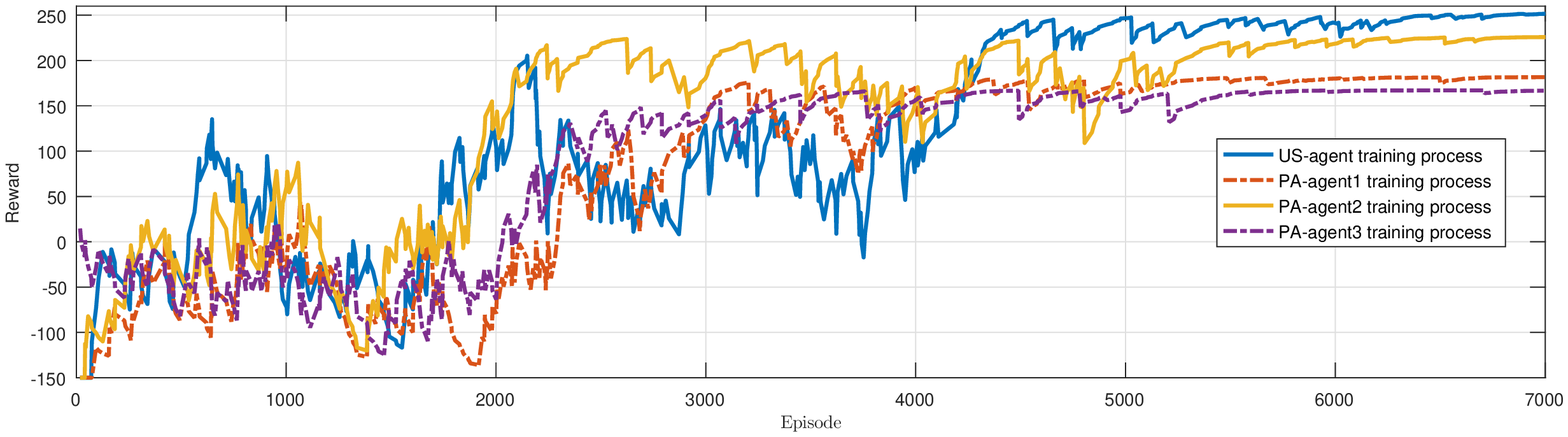}
  }
  \subfigure[]{%\label{fig:subfig1:b} %% label for second subfigure
    \includegraphics[width=4.5in]{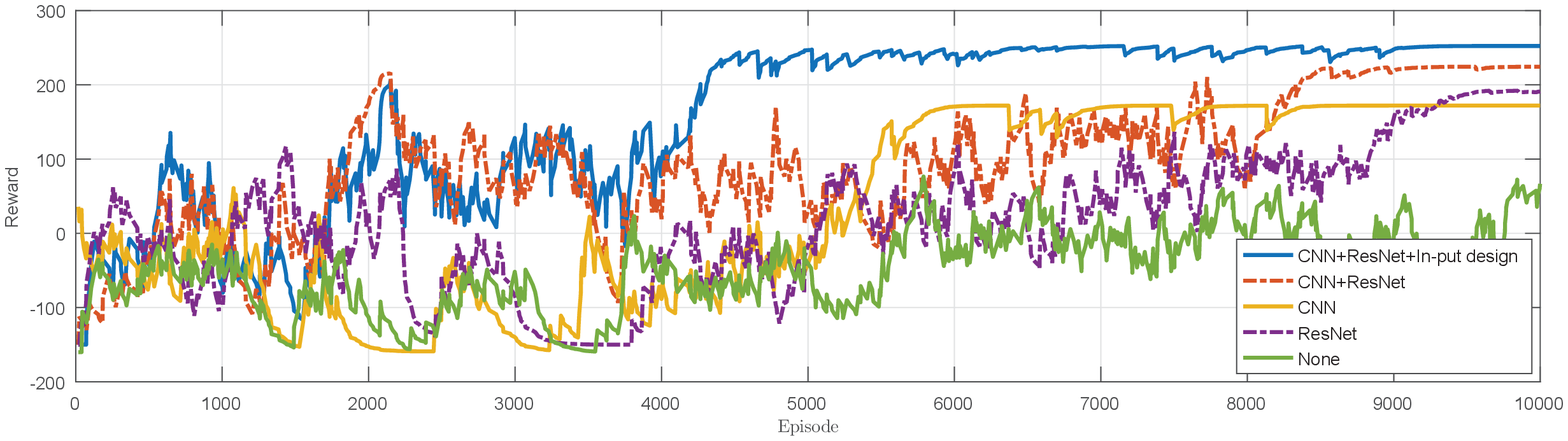}
  }
    \subfigure[]{%\label{fig:subfig1:b} %% label for second subfigure
    \includegraphics[width=4.5in]{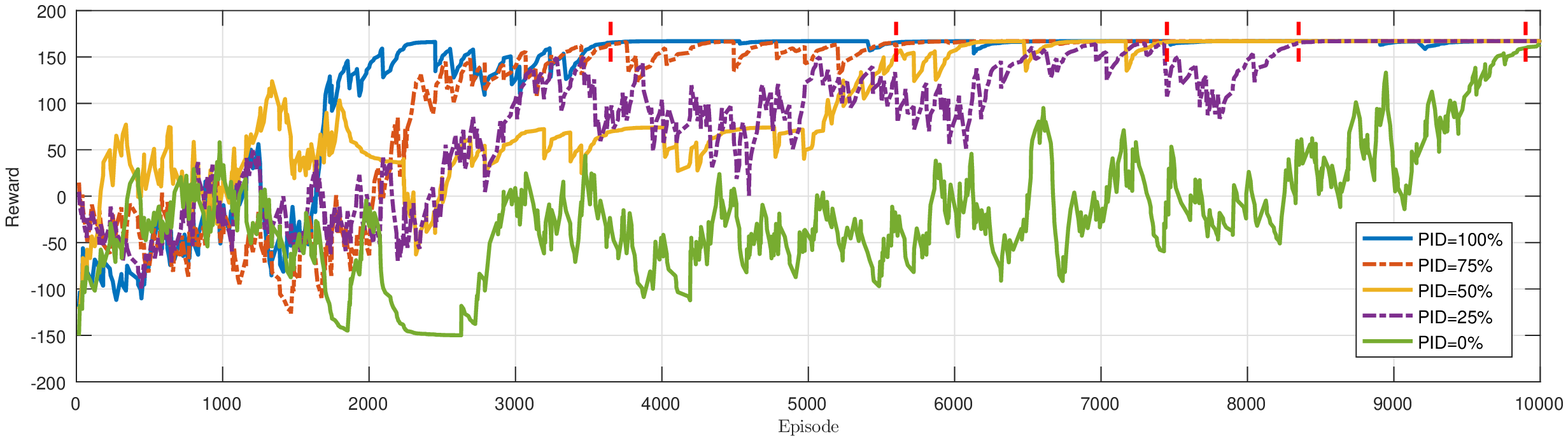}
  }
   \subfigure[]{%\label{fig:subfig1:b} %% label for second subfigure
    \includegraphics[scale=0.47]{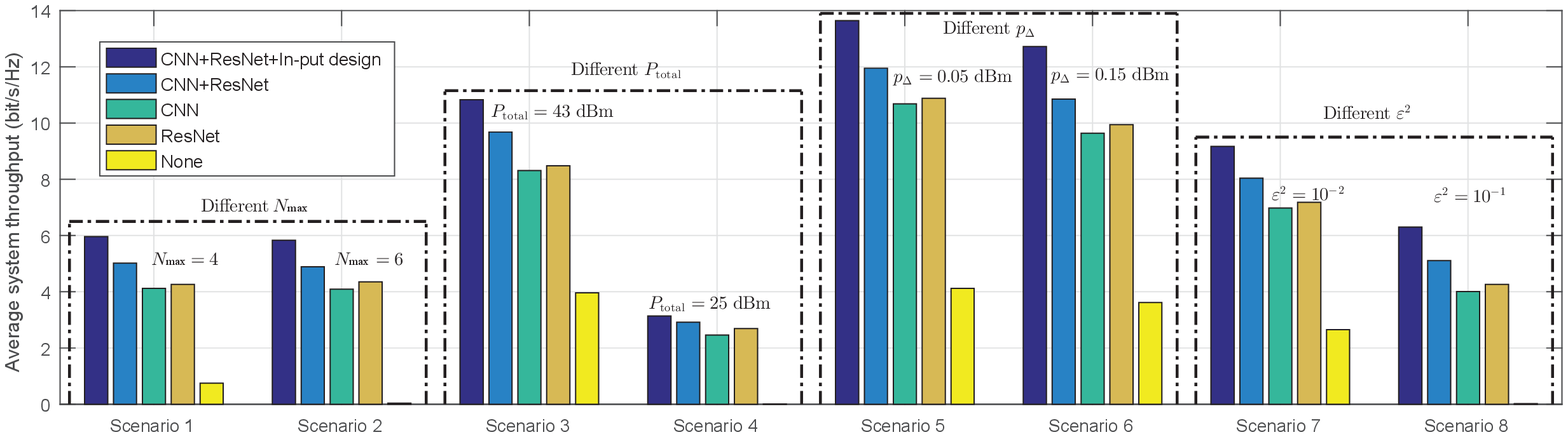}
  }
  \caption{(a) Training process for SA- and PA-agents; (b) Convergence of improvements; (c) Convergence of different PIDs. (d) The impact of different improvements on AT under different scenarios, where 1\&2 $N_{\mathrm{F}}\!=\!20$, $M\!=\!60$, $P_{\mathrm{total}}\!=\!30$ dBm, $p_{\Delta}\!=\!0.05$ dBm, $\varepsilon^{2}\!=\!10^{-4}$; 3\&4 $N_{\mathrm{F}}\!=\!20$, $M\!=\!60$, $N_{\mathrm{max}}\!=\!4$, $p_{\Delta}\!=\!0.03$ dBm, $\varepsilon^{2}\!=\!10^{-3}$; 5\&6 $N_{\mathrm{F}}\!=\!20$, $M\!=\!60$, $N_{\mathrm{max}}\!=\!4$, $P_{\mathrm{total}}\!=\!46$ dBm, $\varepsilon^{2}\!=\!10^{-4}$; 7\&8 $N_{\mathrm{F}}\!=\!15$, $M\!=\!60$, $N_{\mathrm{max}}\!=\!6$, $P_{\mathrm{total}}\!=\!43$ dBm, $p_{\Delta}\!=\!0.05$ dBm.}
  \label{convergence} %% label for entire figure
\end{figure*}
\subsection{Impact of Imperfect SIC}
From Fig. \ref{A_throughput_user_number} to Fig. \ref{A_throughput_power}, we show that JPCA is better than DUCPA, JUSPA, CFJPBA and RUSPA. We compare the AT of DRL-JRM and JPCA in different scenarios. In Fig. \ref{N_change}(a), when $N_{\mathrm{max}}\geq10$, the increase of $N_{\mathrm{max}}$ does not result in a growth of the AT. This is because multiplexing gain by increasing $N_{\mathrm{max}}$ could be offset by the increasingly interference.

In other words, the interference caused by multi-user multiplexing can seriously compromise the performance of MC-NOMA. When $\varepsilon^{2}=10^{-1}$, with the growth of $N_{\mathrm{max}}$, the AT of DRL-JRM and JPCA is increasingly surpassed by that of MC-OMA. Different from what in shown in Fig. \ref{N_change}(a), $P_{\mathrm{total}}$ decreases in Fig. \ref{N_change}(b). The AT of DRL-JRM and JPCA declines under different $\varepsilon^{2}$ values, and the corresponding critical points are moved forward. The same conclusions are drawn from Figs. \ref{N_change}(c) and (d).
\subsection{Convergence and Effectiveness of Different Improvements}
Fig. \ref{convergence}(a) shows the training process of the SA-agent and PA-agents (only three PA-agents are shown due to the limited space). All agents exhibit good convergence. In Fig. \ref{convergence}(b), by adopting ResNet or CNN, the convergence of the network is improved. Although CNN has better convergence than ResNet, its convergent network performance is lower. The input design can also improve the performance and convergence. In Fig.~\ref{convergence}(c), the PID only affects the convergence rate and has no impact on the network performance. Fig. \ref{convergence}(d) shows the AT of the three improvements developed in Section IV-B. The improvements help enhance the throughput and convergence. CNN contributes more to the convergence, while ResNet contributes more to the effectiveness.
\section{Conclusion}
In this paper, we proposed the new DRL-JRM technique for MC-NOMA under hardware sensitivity requirement and imperfect SIC. We evaluated the impact of SIC errors, PDSC, the number of users and the transmit power on resource management, and evaluated the multiplexing capability of MC-NOMA under different interference conditions. Extensive experiments confirmed that DRL-JRM scheme is responsive to different demands of users, and offers good scalability for large-scale problems. DRL-JRM can be potentially extended to allocate contiguous subcarriers (as specified in the 3GPP) by imposing new subcarrier continuity constraints through the output of the ``A'' and ``B'' networks in the SA actor network.

\begin{appendices}
\section{Proof of Theorem 2}
The total time-complexity is given by
\begin{align}
&\resizebox{.88\hsize}{!}{$N_{\mathrm{ep}}[N_{\mathrm{max}}^{\mathrm{SA}}\left(M\left(O\left(\mathrm{SA-CN}\right)+O\left(\mathrm{SA-CN}\right)\right)\right)$
}
\nonumber \\
&\resizebox{.88\hsize}{!}{$+N_{\mathrm{max}}^{\mathrm{PA}}\left(T_{\mathrm{max}}^{\mathrm{PA}}\times\max_{m}\left\{ O\left(\mathrm{state-CNN}_{m}\right)+O\left(\mathrm{SA-CN}_{m}\right)+O\left(\mathrm{SA-CN}_{m}\right)\right\} \right)].
$}
\end{align}
In the PA module, the agents operate in parallel. In our considered scenario, $N_{\mathrm{F}}\!\leq\! M\!\ll\! N_{\mathrm{full}}$. We also approximate the number of neurons $N_{\mathrm{Res}}$ in each layer of ResNet with $N_{\mathrm{full}}$, i.e., $N_{\mathrm{full}}\!\approx\! N_{\mathrm{Res}}$. Then, we have
\begin{align}
&O\left(\mathrm{SA-AN}\right)\approx O\left(2d_{\mathrm{Res}}N_{\mathrm{full}}^{2}+2N_{\mathrm{F}}N_{\mathrm{full}}\right),\label{time_com_1}\\
&O\left(\mathrm{SA-CN}\right)\approx O\left(\left(2d_{\mathrm{Res}}+1\right)N_{\mathrm{full}}^{2}+2N_{\mathrm{F}}N_{\mathrm{full}}\right),\\
&O\left(\mathrm{PA-AN}_{m}\right)\approx O\left(2d_{\mathrm{Res}}N_{\mathrm{full}}^{2}+7N_{\mathrm{F}}N_{\mathrm{full}}\right),\\
&O\left(\mathrm{PA-CN}_{m}\right)\approx O\left(\left(2d_{\mathrm{Res}}+1\right)N_{\mathrm{full}}^{2}+7N_{\mathrm{F}}N_{\mathrm{full}}\right),\\
&\resizebox{.85\hsize}{!}{$O\left(\mathrm{state-CNN}_{m}\right)\approx O\left(\stackrel[l=1]{D}{\sum}\left(M_{l}^{2}K_{l}^{2}C_{l-1}C_{l}\right)+d_{\mathrm{cnn}}N_{\mathrm{full}}^{2}\right). $}
\label{time_com_2}
\end{align}
Based on (\ref{time_com_1})--(\ref{time_com_2}), the time-complexity of the SA module is
\begin{align}\label{time_com_total_1}
\resizebox{.85\hsize}{!}{$O_{1}\left(N_{\mathrm{ep}}N_{\mathrm{max}}^{\mathrm{SA}}N_{\mathrm{full}}M\left(4d_{\mathrm{Res}}N_{\mathrm{full}}+N_{\mathrm{full}}+4N_{\mathrm{F}}\right)\right),
$}
\end{align}
and the time-complexity of the PA module is given by
\begin{align}\label{time_com_total_2}
\resizebox{.85\hsize}{!}{$O_{2}\left(N_{\mathrm{ep}}N_{\mathrm{max}}^{\mathrm{PA}}T_{\mathrm{max}}^{\mathrm{PA}}\left(\stackrel[l=1]{D}{\sum}\left(M_{l}^{2}K_{l}^{2}C_{l-1}C_{l}\right)+\left(d_{\mathrm{net}}N_{\mathrm{full}}+14N_{\mathrm{F}}\right)N_{\mathrm{full}}\right)\right),
$}
\end{align}
where $d_{\mathrm{net}}=4d_{\mathrm{Res}}+1+d_{\mathrm{cnn}}$. The overall time-complexity sums up (\ref{time_com_total_1}) and (\ref{time_com_total_2}).

The total space-complexity is given by
 \begin{align}
&O\left(\mathrm{SA-AN}\right)+O\left(\mathrm{SA-CN}\right) +O\left(\mathrm{pool}\right)\\ \nonumber
&\resizebox{.85\hsize}{!}{$+M\times\left(O\left(\mathrm{state-CNN}_{m}\right)+O\left(\mathrm{PA-AN}_{m}\right)+O\left(\mathrm{PA-CN}_{m}\right)\right).
$}
\end{align}
where $O\left(\mathrm{pool}\right)$ is the space-complexity of the ``experience replay pool''. Compared with the ``experience replay pool'', the space-complexity of the state and action spaces is comparatively negligible.
With reference to the time-complexity, the space-complexity of the SA module is
 \begin{align}\label{space_com_total_1}
O_{1}\left(N_{\mathrm{full}}\left(4d_{\mathrm{Res}}N_{\mathrm{full}}+N_{\mathrm{full}}+4N_{\mathrm{F}}\right)\right).
\end{align}
The space-complexity of the PA module is given by
 \begin{align}
 \resizebox{.85\hsize}{!}{$O_{2}\left(\stackrel[l=1]{D}{\sum}\left(K_{l}^{2}C_{l-1}C_{l}\right)+\stackrel[l=1]{D}{\sum}\left(M^{2}C_{l}\right)+\left(d_{\mathrm{net}}N_{\mathrm{full}}+14N_{\mathrm{F}}\right)N_{\mathrm{full}}\right).
$}
\end{align}
The space-complexity of the ``experience replay pool'' is
 \begin{align}\label{space_com_total_3}
 &\resizebox{.88\hsize}{!}{$O_{3}\left(\mathrm{pool}\right)=O_{3}\left(EM^{\mathrm{u}}\times\left(2\left(2N_{\mathrm{F}}+2\right)+N_{\mathrm{F}}+1\right)+M\cdot EM_{m}^{\mathrm{p}}\times\left(2\left(6N_{\mathrm{F}}+4\right)+N_{\mathrm{F}}+1\right)\right)$}
\nonumber \\
&\approx O_{3}\left(18M\cdot N_{\mathrm{F}}\cdot EM^{\mathrm{u}}\right),
\end{align}
where $EM^{\mathrm{u}}\approx EM_{m}^{\mathrm{p}}$. The overall space-complexity is the sum of (\ref{space_com_total_1})--(\ref{space_com_total_3}).
\qed
\end{appendices}

\bibliography{ciations}
\begin{IEEEbiography}[{\includegraphics[width=1in,height=1.25in,clip,keepaspectratio]{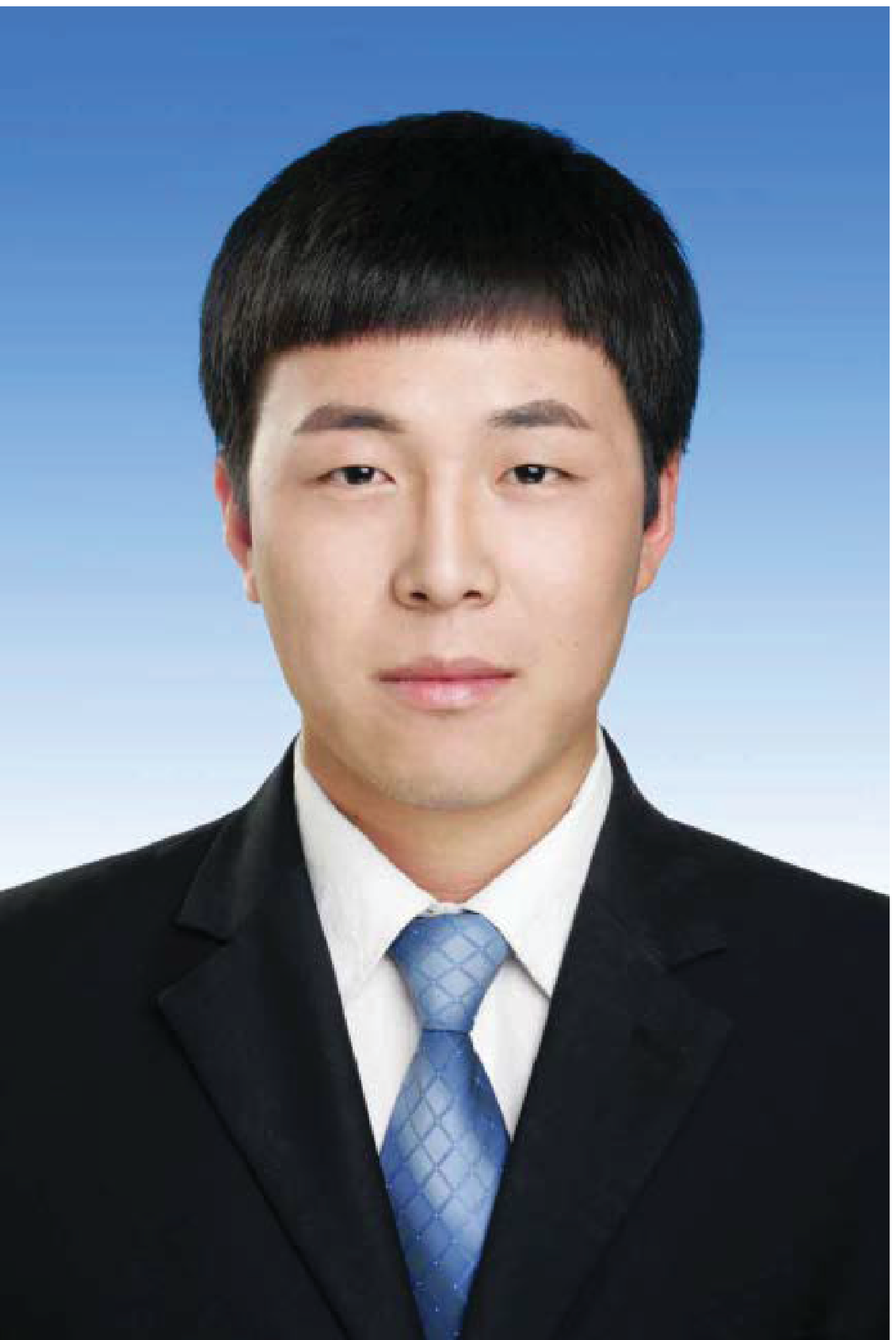}}]{Shaoyang Wang}
(S'18) received the B.E. degree in electronic information science and technology from Shandong University, China, in 2017. He is currently pursuing the Ph.D. degree with the School of Information and Communication Engineering, Beijing University of Posts and Telecommunications (BUPT), Beijing, China. His current research interests include non-orthogonal multiple access access, network function virtualization, and intelligent wireless resource management.
\end{IEEEbiography}
\begin{IEEEbiography}[{\includegraphics[width=1in,height=1.25in,clip,keepaspectratio]{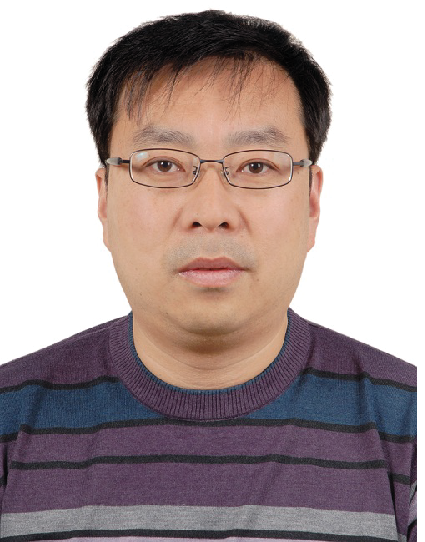}}]{Tiejun Lv}
(M'08-SM'12) received the M.S. and Ph.D. degrees in electronic engineering from the University of Electronic Science and Technology of China (UESTC), Chengdu, China, in 1997 and 2000, respectively. From January 2001 to January 2003, he was a Postdoctoral Fellow with Tsinghua University, Beijing, China. In 2005, he was promoted to a Full Professor with the School of Information and Communication Engineering, Beijing University of Posts and Telecommunications (BUPT). From September 2008 to March 2009, he was a Visiting Professor with the Department of Electrical Engineering, Stanford University, Stanford, CA, USA. He is the author of three books, more than 90 published IEEE journal papers and 200 conference papers on the physical layer of wireless mobile communications. His current research interests include signal processing, communications theory and networking. He was the recipient of the Program for New Century Excellent Talents in University Award from the Ministry of Education, China, in 2006. He received the Nature Science Award in the Ministry of Education of China for the hierarchical cooperative communication theory and technologies in 2015.
\end{IEEEbiography}
\begin{IEEEbiography}[{\includegraphics[width=1in,height =1.25in,clip,keepaspectratio]{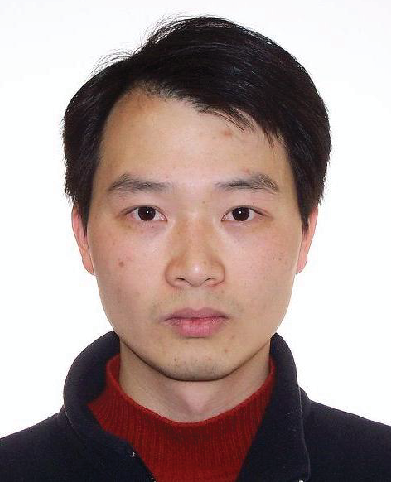}}]{Wei Ni} (M'09-SM'15) received the B.E. and Ph.D. degrees in Electronic Engineering from Fudan University, Shanghai, China, in 2000 and 2005, respectively. Currently, he is a Group Leader and Principal Research Scientist at CSIRO, Sydney, Australia, and an Adjunct Professor at the University of Technology Sydney and Honorary Professor at Macquarie University, Sydney. He was a Postdoctoral Research Fellow at Shanghai Jiaotong University from 2005 to 2008; Deputy Project Manager at the Bell Labs, Alcatel/Alcatel-Lucent from 2005 to 2008; and Senior Researcher at Devices R\&D, Nokia from 2008 to 2009. His research interests include signal processing, optimization, learning, and their applications to network efficiency and integrity.
\end{IEEEbiography}
\begin{IEEEbiography}[{\includegraphics[width=1in,height=1.25in,clip,keepaspectratio]{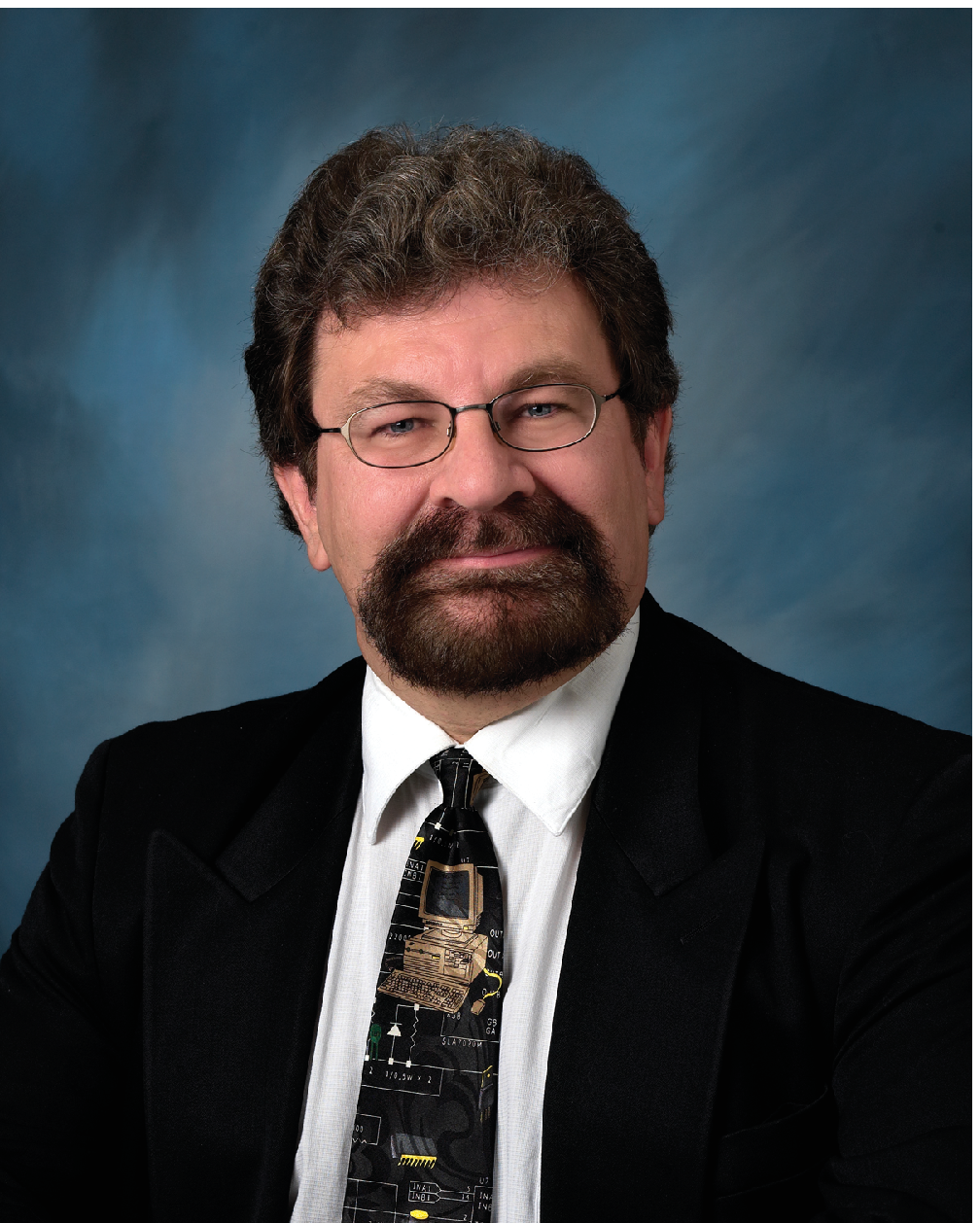}}]{Norman C. Beaulieu} is an Academician of the Royal Society of Canada (RSC), and an Academician of the Canadian Academy of Engineering (CAE). He is a Fellow of the Institute of Electrical and Electronics Engineers (IEEE), a Fellow of the Institution of Engineering and Technology (IET) of the United Kingdom, a Fellow of the Engineering Institute of Canada (EIC), and a Nicola Copernicus Fellow of Italy. He is the recipient of the esteemed Natural Sciences and Engineering Research Council (NSERC) of Canada E.W.R. Steacy Memorial Fellowship. He is the only person in the world to hold both the IEEE Edwin Howard Armstrong Award and the IEEE Reginald Aubrey Fessenden Award, named for the inventors of Frequency Modulation or FM, and Amplitude Modulation or AM, respectively. Prof. Beaulieu is a Beijing University of Posts and Telecommunications BUPT Thousand-Talents Scholar. He holds the third highest Web of Science ISI h-index in the world in the combined areas of communication theory and information theory. Prof. Beaulieu was awarded the title ``State Especially Recruited   Foreign Expert'' certified upon him by Minister of Human Resources and Social Insurance, and Vice Minister of the Organization Department, Yi Weimin. Pro. Beaulieu is the recipient of the Royal Society of Canada Thomas W. Eadie Medal, the Medaille K.Y. Lo Medal of the EIC, and was the subject of a TIME Magazine feature article.  He was also awarded the unique Special University Prize in Applied Science of the University of British Columbia, and the J. Gordin Kaplan Award for Research of the University of Alberta.
\end{IEEEbiography}
\begin{IEEEbiography}[{\includegraphics[width=1in,height=1.25in,clip,keepaspectratio]{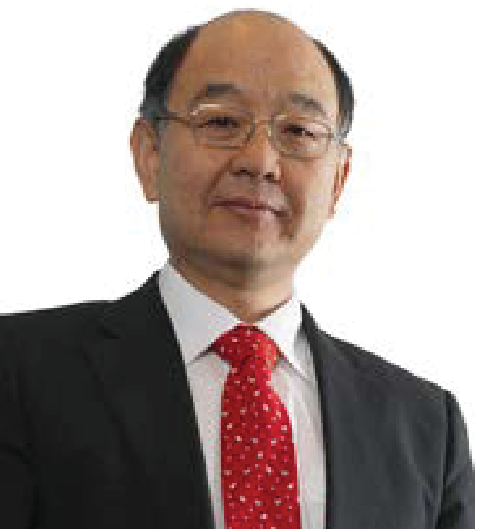}}]{Y. Jay Guo}
(Fellow'2014) received a Bachelor Degree and a Master Degree from Xidian University in 1982 and 1984, respectively, and a PhD Degree from Xian Jiaotong University in 1987, all in China. His research interest includes antennas, mm-wave and THz communications and sensing systems as well as big data technologies. He has published four books and over 550 research papers including 280 journal papers, most of which are in IEEE Transactions, and he holds 26 patents. He is a Fellow of the Australian Academy of Engineering and Technology, a Fellow of IEEE and a Fellow of IET, and was a member of the College of Experts of Australian Research Council (ARC, 2016-2018). He has won a number of most prestigious Australian Engineering Excellence Awards (2007, 2012) and CSIRO Chairman's Medal (2007, 2012). He was named one of the most influential engineers in Australia in 2014 and 2015, respectively, and one of the top researchers in Australia in 2020.

He is a Distinguished Professor and the Director of Global Big Data Technologies Centre (GBDTC) at the University of Technology Sydney (UTS), Australia. Prior to this appointment in 2014, he served as a Director in CSIRO for over nine years. Before joining CSIRO, he held various senior technology leadership positions in Fujitsu, Siemens and NEC in the U.K.

Prof Guo has chaired numerous international conferences and served as guest editors for a number of IEEE publications. He is the Chair of International Steering Committee, International Symposium on Antennas and Propagation (ISAP). He was the International Advisory Committee Chair of IEEE VTC2017, General Chair of ISAP2022, ISAP2015, iWAT2014 and WPMC'2014, and TPC Chair of 2010 IEEE WCNC, and 2012 and 2007 IEEE ISCIT. He served as Guest Editor of special issues on ``Low-Cost Wide-Angle Beam Scanning Antennas'', ``Antennas for Satellite Communications'' and  ``Antennas and Propagation Aspects of 60-90GHz Wireless Communications,'' all in IEEE Transactions on Antennas and Propagation, Special Issue on ``Communications Challenges and Dynamics for Unmanned Autonomous Vehicles,'' IEEE Journal on Selected Areas in Communications (JSAC), and Special Issue on ``5G for Mission Critical Machine Communications'', IEEE Network Magazine.
\end{IEEEbiography}
\end{document}